\documentclass[review]{elsarticle}

\journal{Journal of \LaTeX\ Templates}

\usepackage{epsf,amsmath,amssymb,graphicx,epsfig}
\usepackage[ruled]{algorithm2e}
\usepackage{caption}
\usepackage{subfigure}
\usepackage{xr}
\usepackage[colorlinks]{hyperref}
\begin{document}

\begin{frontmatter}

\title{Continuous Deep Hierarchical Reinforcement Learning for Ground-Air Swarm Shepherding}

\author{Hung The Nguyen, Tung Duy Nguyen,Vu Phi Tran,
Matthew Garratt,Kathryn Kasmarik, Sreenatha Anavatti,
Michael Barlow, and~Hussein A. Abbass}
\address{School of Engineering and Information Technology, University of New South Wales, Canberra, Australia. \\ E-mail: \{hung.nguyen,tung.nguyen\}@student.adfa.edu.au; \{phi.tran, k.kasmarik, m.garratt, s.anavatti, m.barlow,h.abbass\}@adfa.edu.au}


\begin{abstract}
The control and guidance of multi-robots (swarm) is a non-trivial problem due to the complexity inherent in the coupled interaction among the group. Whether the swarm is cooperative or non-cooperative, lessons can be learnt from sheepdogs herding sheep. Biomimicry of shepherding offers computational methods for swarm control with the potential to generalize and scale in different environments. However, learning to shepherd is complex due to the large search space that a machine learner is faced with. We present a deep hierarchical reinforcement learning approach for shepherding, whereby an unmanned aerial vehicle (UAV) learns to act as an aerial sheepdog to control and guide a swarm of unmanned ground vehicles (UGVs). The approach extends our previous work on machine education to decompose the search space into a hierarchically organized curriculum. Each lesson in the curriculum is learnt by a deep reinforcement learning model. The hierarchy is formed by fusing the outputs of the model. The approach is demonstrated first in a high-fidelity robotic-operating-system (ROS)-based simulation environment, then with physical UGVs and a UAV in an in-door testing facility. We investigate the ability of the method to generalize as the models move from simulation to the real-world and as the models move from one scale to another.
\end{abstract}

\begin{keyword}
Deep Reinforcement Learning, Machine Education, Shepherding, Swarm Robotics, Unmanned Aerial Vehicles, Unmanned Ground Vehicles
\end{keyword}

\end{frontmatter}


\section{Introduction}
A wide variety of attempts have been made to design computational intelligence algorithms to design swarm robotic systems~\cite{martinez2007motion,Chang2012Fuzzy}. One ultimate objective has been centered around the concept of emergence, whereby a long standing question is how to design local rules that create self-organized and interesting group-level emergent properties~\cite{carelli2006centralized,Chang2012Fuzzy,oh2017bio}. The practical motivation is to design multi-robot systems that can cooperatively solve problems that a single robot, on its own, cannot~\cite{wen2018swarm,yan2019efficient}.

A swarm of multi-robot systems demands higher computational requirements as the swarm size increases. We hypothesize that the scalability problem could be solved by borrowing concepts from Nature, where a single agent could control a flock. The computational shepherding problem~\cite{long2019comprehensive}, inspired from real shepherding in agriculture, is borrowed in this paper for swarm control.

Similar to a number of other authors, Str\"{o}mbom et al.~\cite{strombom2014solving} developed a heuristic model and validated it against actual sheepdog behaviors. An artificial agent mimicked the sheepdog and a swarm of artificial agents mimicked the behavior of the sheep. The flexibility of the model could extend its uses to human-swarm interaction and to the dynamic control of a robotic swarm using a single, or a few, control agents. A large swarm could be guided and controlled using a smaller number of well-trained artificial sheepdogs. 

Swarm guidance in the form of shepherding, offers an opportunity for an agent in one domain to guide a swarm in a different domain. For example, the coordination between unmanned aerial vehicles (UAVs) and unmanned ground vehicles (UGVs) has been investigated in previous studies for search-and-rescue, surveillance, path planning for delivery, and wildlife research missions~\cite{Chen2016coordination,Minaeian2016autonomous,mathew2015planning,mathews2019supervised}.
UAVs possess an aerial mode of observation which provides an extensive field of view (FoV); they have significant advantages when guiding ground vehicles~\cite{nguyen2017supervised} and, therefore, can plan and guide motion from the air. It is plausible to use shepherding principles to design an effective coordination strategy or autonomous air-ground vehicles~\cite{chaimowicz2004aerial}. The swarm control problem in this case aims at developing an artificial intelligence for controlling a UAV to influence the ground vehicles. The coupling in the dynamics creates a repulsive force from the UGVs due to the presence of the UAV.

By leveraging the generalization ability of a deep neural network and the long-term planning ability of reinforcement learning algorithms, deep reinforcement learning methods~\cite{mnih2016asynchronous,yang2018hierarchical,yasuda2018collective} could potentially design an intelligent agent which controls the UAV in the proposed aerial shepherding problem. Nevertheless, it is challenging when one designs a deep reinforcement learning algorithm for this problem due to the complexity of the search space and uncertainties inherent in the operation of real vehicles and the environment~\cite{nguyen2017supervised}. Previous studies of machine education~\cite{gee2019transparent,clayton2019machine} have demonstrated the feasibility of reducing the complexity of the learning problem by disassembling the search space into smaller chunks.

Our previous research~\cite{nguyen2019deep} applied a decomposition approach using deep hierarchical reinforcement learning (DHRL). The approach was successful in designing a UAV-shepherd to guide a group of ground vehicles. The performance of the DHRL is competitive with the rule-based baseline method of the Str\"{o}mbom method~\cite{strombom2014solving}. In this paper, we advance the research in three directions. 

Firstly, we use a Hierarchical Deep Deterministic Policy Gradients (H-DDPGs) methodology to design continuous control policies. Using the same aerial shepherding task as conducted in our DHRL approach~\cite{nguyen2019deep}, we train an aerial sheepdog to guide a swarm of UGVs, and decompose the entire mission into chunks. Each chunk corresponds to a sub-task that requires learning of a group of basic sub-skills. In the context of shepherding, two basic sub-skill groups include the ability to collect the sheep into a flock and the ability to drive the flock to a goal. Each of these sub-skills is learnt individually using DDPGs then the sub-skills are aggregated to solve the overall problem using H-DDPGs. When we compare the baseline method of Str\"{o}mbom and our previous DHRL work against the proposed H-DDPG, the proposed method is at least as good in terms of performance, with better success rate and efficiency.

Secondly, we transfer the models learnt in a simulated environment to real UAV-UGVs in our indoor testing facility. We show that the UAV driven by H-DDPGs is able to perform effectively in this system. 

Thirdly, we validat the robustness of transferring the models trained in a small environment to larger environments. We train an H-DDPG model in a small simulation environment in order to reduce training time. We then successfully transfer the model to a larger simulation environment.

The remainder of the paper is organized as follows. In Section~\ref{section-2}, we formally define shepherding using an appropriate notational system and a corresponding mathematical objective.  The proposed H-DDPG framework is introduced in Section~\ref{section-3}, along with the decomposition algorithm, reward scheme, and testing method. The framework is then applied to a physical UAV-UGV shepherding task in Section~\ref{section-4}. Sections~\ref{section-5} and~\ref{section-6} present the results of the framework in simulation and physical environments, respectively. Conclusions are drawn in Section~\ref{section-7}, followed by a discussion on future work.

\section{Related work}

Two approaches have emerged for the control of a robotic swarm: rule-based and learning-based techniques~\cite{oh2015survey,balch1998behavior}. Rule-based systems~\cite{vasarhelyi2014outdoor,Ramazani2017Rigidity} represent the controller in different forms including symbolic if...then..., tree representations, predicate or higher-order logic, equations using calculus-based or probabilistic representations,  or graph-representations such as finite state machines and probabilistic graphical models. These systems define and fix the mapping from the states of an agent into action vectors~\cite{sen2017cooperative,guillet2017formation,miao2018distributed}.
Learning systems dynamically form a model from experiences and interactions with the environment~\cite{Hung2017qlearning}.

Rule-based systems are scalable and fast to adapt in the contexts they were designed for. However, when the context changes, as in changes in the distribution of uncertainties and disturbances~\cite{yi2017bio}, such that the model is no longer appropriate, they are unable to adapt or generalize to new contexts. Learning systems, however, are flexible, so that they can adapt to control a unmanned vehicle or swarm in novel situations~\cite{santoso2017state}. 

Literature on other swarm behavior learning problems such as swarm flocking and leader-follower models have used decentralized policies learned with reinforcement learning (RL)~\cite{Hung2017qlearning,Yang2018leader}. A reward scheme is used in RL to dynamically design a learning model for intelligent lifelong learning agents able to adapt to changing environments using trial and error~\cite{singh2016navigation}. Asada et al~\cite{asada1999cooperative} uses RL to learn cooperative behaviors in a robotics soccer application.
Zema et al~\cite{Zema2019formation} introduces a Q-learning algorithm, applied on each UAV follower, which uses the radio signal strength values from communications among followers and a single leader to learn to maintain the swarm\textquoteright s formation. 
Coupled with reinforcement learning methods for producing control policy, knowledge sharing techniques are commonly used in swarm systems to distribute information on the environment or a common value system among swarm members~\cite{nguyen2018swarm,de2019bio,palmer2018lenient}. 

\section{Problem Formulation}\label{section-2}

Without loss of generality, we assume the environment is squared of length $L$, with two types of agents forming two sets: a set of $N$ sheep $\Pi = \{ \pi_1, \dots, \pi_i, \dots, \pi_N \}$, and a set of $M$ shepherds $B = \{ \beta_1, \dots, \beta_j, \dots, \beta_M \}$. A sheep $\pi_i$ performs three basic behaviors at a time step $t$:

\begin{itemize}
\item \emph{Escaping behavior} $\sigma_3$: Sheep $\pi_i$ at position $P^t_{\pi_i}$ attempts to escape a predator (sheepdog) by a repulsive force $F^t_{\pi_i\beta_j}$ if the distance between the sheep and sheepdog $\beta_j$ is less than the sheep sensing range for sheepdog $R_{\pi\beta}$ (Equation~\ref{eq:SheepDistanceToShepherdEquation}). 

\begin{equation}\label{eq:SheepDistanceToShepherdEquation} \|P^t_{\pi_i}-P^t_{\beta_j}\| \le R_{\pi\beta} \end{equation}

The force vector of this behavior is computed by:

\begin{equation}\label{eq:SheepEscapeEquation} F^t_{\pi_i\beta} = \sum_j \frac{P^t_{\pi_i}-P^t_{\beta_j}}{\|P^t_{\pi_i}-P^t_{\beta_j}\|}, \forall j \ where \ \|P^t_{\pi_i}-P^t_{\beta_j}\| \le R_{\pi\beta} \end{equation}

\item \emph{Collision avoidance} $\sigma_4$: Sheep $\pi_i$ avoids collision with another sheep $\pi_{k\ne i}$ using a force vector $F^t_{\pi_i\pi_{-i}}$ representing the summation of all repulsive force vectors from all sheep within the neighborhood. Behavior $\sigma_5$ exists when at least the distance between one other sheep and sheep $\pi_i$ is less than $R_{\pi\pi}$; that is,
 
\begin{equation}\label{eq:SheepDistanceToOtherSheepEquation} \exists {k}, \ such \ that \ \|P^t_{\pi_i}-P^t_{\pi_{k}}\| \le R_{\pi\pi} \end{equation}

The force vector for this behavior is then computed by:

\begin{equation}\label{eq:SheepCollisionAvoidanceEquation} F^t_{\pi_i\pi_{i}} = \sum_{i1} \frac{P^t_{\pi_i}-P^t_{\pi_{i1}}}{\|P^t_{\pi_i}-P^t_{\pi_{i1}}\|}  \forall {i1} \ where \ \|P^t_{\pi_i}-P^t_{\pi_{i1}}\| \le R_{\pi\pi} \end{equation}

\item \emph{Grouping behavior} $\sigma_5$: Sheep $\pi_i$ gets attracted to the local center of mass of the flock in its neighborhood $\Lambda^t_{\pi_i}$ with a force $F^t_{\pi_i\Lambda^t_{\pi_i}}$:
\begin{equation}\label{eq:SheepGroupingEquation} F^t_{\pi_i\Lambda^t_{\pi_i}} = \frac{\Lambda^t_{\pi_i} - P^t_{\pi_i} }{\|\Lambda^t_{\pi_i} - P^t_{\pi_i} \|} \end{equation}

\end{itemize}

The effect of total force vector from previous time step $F^{t-1}_{\pi_i}$ with weight $W_{\pi_\upsilon}$ is also taken into account. Finally, the weighted summation of all force vectors $F^t_{\pi_i}$ is computed to determine the movement of sheep $\pi_i$.

\begin{equation}\label{eq:SheepTotalForceEquation}
F^t_{\pi_i}  = W_{\pi_\upsilon}  F^{t-1}_{\pi_i} +
W_{\pi \Lambda} F^t_{\pi_i{\Lambda^t_{\pi_i}}} + W_{\pi\beta} F^t_{\pi_i\beta_j} + W_{\pi\pi}  F^t_{\pi_i\pi_{-i}}
\end{equation}

Str\"{o}mbom et al model specifies two basic behaviors for a shepherd $\beta_j$:
\begin{itemize}
\item \emph{Driving} behavior $\sigma_1$: The behavior is triggered when distances of all sheep to their center of mass is smaller than a threshold $f(N)$ specified in Equation~\ref{eq:ShepherdBehaviorSelectionThresholdEquation}; that is, all sheep are gathered as a cluster within a circle of small enough diameter. The shepherd $\beta_j$ at position $P^t_{\beta_j}$ moves according to a force vector $F^t_{\beta_jcd}$. The direction of the vector emits from the shepherd $\beta_j$ to a driving sub-goal point $P^t_{\beta_j \sigma_1}$ situated a distance of $R_{\pi\pi} \sqrt{N}u$ behind the sheep\textquoteright s cluster relative to the goal location. $R_{\pi\pi}$ is the sensing range between two sheep, and $u$ is the unit distance. The driving force vector is computed by Equation~\ref{eq:ShepherdDrivingForceEquation}.
\begin{equation}\label{eq:ShepherdBehaviorSelectionThresholdEquation} f(N) = R_{\pi\pi} N^\frac{2}{3}\end{equation}
\begin{equation}\label{eq:ShepherdDrivingForceEquation} F^t_{\beta_jcd} = \frac{P^t_{\beta_j\sigma_1} - P^t_{\beta_j}}{\|P^t_{\beta_j\sigma_1} - P^t_{\beta_j} \|}
\end{equation}

\item \emph{Collecting behavior} $\sigma_2$: If there is at least one sheep outside the flock allowed radius $f(N)$; that is, there exists sheep that need to be brought back to the sheep flock, the shepherd $\beta_j$ moves according to a force vector $F^t_{\beta_jcd}$. The vector emits from the shepherd $\beta_j$ the collection point $P^t_{\beta_j \sigma_2}$ situated a distance of $R_{\pi\pi} u$ behind the furthest sheep relative to the flock center of mass. The collecting force vector is computed by Equation~\ref{eq:ShepherdCollectionForceEquation}.
\begin{equation}\label{eq:ShepherdCollectionForceEquation} F^t_{\beta_jcd} = \frac{P^t_{\beta_j\sigma_2} - P^t_{\beta_j}}{\|P^t_{\beta_j\sigma_2} - P^t_{\beta_j} \|}\end{equation}
\end{itemize}

The updated positions of each agent are computed by Equations~\ref{eq:UpdatedShepherdPositionEquation} and~\ref{eq:UpdatedSheepPositionEquation}, where $S^t_{\beta_j}$ and $S^t_{\pi_i}$ are the speed at time $t$ of shepherd $\beta_j$ and sheep $\pi_i$, respectively. \\
\begin{equation}\label{eq:UpdatedShepherdPositionEquation}
P^{t+1}_{\beta_j} = P^{t}_{\beta_j} + S^t_{\beta_j} F^t_{\beta_j}
\end{equation}
\begin{equation}\label{eq:UpdatedSheepPositionEquation}
P^{t+1}_{\pi_i} = P^t_{\pi_i} + S^t_{\pi_i} F^t_{\pi_i}
\end{equation}  

There are two main objectives in the shepherding problem. Firstly, the shepherd agents need to collect the sheep from a scattered swarm into a flock and herd them towards a goal destination. The first objective is to find a solution that minimizes the completion time. Given $T$ as the completion time for the shepherding task, the course of actions taken by the shepherds should result in the minimum time possible to complete and the state of the sheep flock must satisfy the constraints below:
\begin{equation}
\begin{split}
T^* = min(T) \; \text{such that} \\      
 \begin{cases} \forall{\pi_i \in \Pi}, \;\;\; ||P^T_{\pi_i}-P^T_G|| \leq \mathbb{D} \\ \forall{\pi_i \in \Pi}, \forall{\beta_j \in B}, \;\;\; ||\Lambda^T_{\beta_j} - P^T_{\pi_i}|| \leq f(N) \end{cases}
\end{split}
\end{equation}

where $P^T_G$ is the position of the goal at time $T$ and $\Lambda^T_{\beta_j}$ is the local centre of mass of the shepherd $j$.

In addition, the second objective is to minimize the total travel distance of the shepherds, that is
\begin{equation}
\Delta \mathbf{P^*_{\beta_j}} = min \sum_{t=1}^T\sum_{j=1}^M ||P^{t+1}_{\beta_j} - P^t_{\beta_j}||
\end{equation}

\section{Deep Hierarchical Reinforcement Learning}\label{section-3}

Our previous study~\cite{nguyen2019deep} has demonstrated that a deep hierarchical reinforcement learning (DHRL) framework with multi-skill learning is capable of producing an effective strategy for aerial shepherding of ground vehicles. Simulation results of the study indicate that the performance of the DHRL is equivalent to the performance of the Str\"{o}mbom model. A limitation of DHRL is its reliance on Deep Q-Networks which only generate four discrete actions: moving forward, backward, turning right, and turning left, corresponding to four axis-parallel movements with fixed travel distance every step. In reality, the action space of a UAV is continuous, i.e the UAV moves according to an arbitrary vector with dynamic length. Hence, the trajectory produced by the DHRL might be sub-optimal and less smooth than a continuous output policy.   

To address that limitation, we propose a Hierarchical Deep Deterministic Policy Gradient (H-DDPG) algorithm for aerial shepherding of autonomous rule-based UGV agents. The algorithm inherits the advantages of a deep hierarchical reinforcement learning framework, but combines with a continuous action-producing type of reinforcement learning networks. The H-DDPG produces a continuous output policy for the UAV. We first describe the DDPG algorithm for controlling the UAV in Subsection~\ref{section-3.1} then introduce our proposed learning framework for multi-skill learning in shepherding application in subsection~\ref{section-3.2}. In Section~\ref{section-5}, we also compared our proposed H-DDPG approach with the DHRL and the Str\"{o}mbom method~\cite{strombom2014solving} in a simulation environment.

\subsection{Deep Deterministic Policy Gradient/DDPG}\label{section-3.1}

\emph{Deep Reinforcement Learning/DRL} (DRL)~\cite{mnih2015human} couples a multi-layer hierarchy of deep neural networks with  optimal planning using reinforcement learning for effective behavioral learning in large and continuous state spaces. Less emphasis is placed on feature engineering due to the ability of deep models to autonomously approximate necessary features.

This section revisits the formulation of the reinforcement learning (RL) problem and then describes the Deep Deterministic Policy Gradient (DDPG) algorithm. RL searches for the strategy that offers the best long-term outcome at each state of the environment. Consider a state $s_t$ at time $t$, an agent can perform one action $a_t$ which leads the agent to a new state $s_{t+1}$ in the state space. An instant reward $w_t$ corresponding to the return of the performed action and the value of the next state is received by the agent. The objective is to find the best strategy, given the states and the reward function, with the maximum accumulated return over time:
\begin{equation}\label{eq:3}
W_t = \sum^T_{t'=t}\gamma^{t'-t}w_{t'}
\end{equation}
where $T$ is the total number of time steps, and $\gamma$ is the discount factor.

Model-free approaches in RL consider the problem in the absence of a world model which are highly suitable for real problems in novel and unknown environments. In these approaches, the Q-value $Q(s_t,a_t)$ represents the expected value of each state-action pair. Instead of storing Q-values for every state-action pairs, which is infeasible in continuous environments, DRL uses deep neural networks as universal functional approximators to approximate the mapping from input states to Q-values.

While many problems have been solved using Deep Q-networks (DQN), a popular DRL method in the literature is DDPG~\cite{lillicrap2015continuous}, which outputs continuous actions. The method is more appropriate for shepherding to learn and generate the influence vectors produced by sheepdogs in the environment.

DDPG employs an architecture that involves two networks called actor and critic. The actor network (with weights $\theta ^{\tau}$) estimates the output policy in the form of continuous values. To approximate the corresponding Q-value of the policy proposed by the actor, the critic network (with weights $\theta ^{Q}$) is used. Common practice of the DDPG algorithm is to initialize those two main networks and their clones called target networks (with weights $\theta ^{\tau^\prime}$ and $\theta ^{Q^\prime} $ respectively) in order to stabilize the learning process that happens when the networks change too quickly. The loss function for the critic network is shown below:
\begin{equation} 
\zeta(\theta ^{Q})={ (Q (s_{t},a_{t} | \theta ^{Q}) -y_{t})}^{2} 
\end{equation}
where the target value $y_{t}$ is computed as followings:
\begin{equation}\label{eq:ComputeTargetValue} 
y_{t}=w_{t}+\gamma Q(s_{t+1}, \tau  (s_{t+1} | \theta ^{\tau^\prime })| \theta ^{Q^\prime }) 
\end{equation}

The updating process of the deep critic network is based on the following equation:
\begin{equation}\label{eq:UpdateCriticNetwork}
\theta ^{Q}\leftarrow \theta^{Q} - \alpha _{Q} \nabla _{\theta ^{Q}}\zeta(\theta ^{Q}) \end{equation}

where $\alpha_{Q}$ is the learning rate of the critic network. The action gradient computed with the critic network is then used for updating the weights of the actor network $\theta ^{\tau }$:
\begin{equation}\label{eq:UpdateActorNetwork} 
\theta ^{\tau }\leftarrow \theta ^{\tau } - \alpha _{\tau } \nabla _{a}Q ({s_{t},\tau  (s_{t}|\theta ^{\tau }) }| \theta ^{Q}) \nabla _{\theta ^{\tau }}\pi  (s_{t}\vert \theta ^{\tau }) 
\end{equation}
where $\alpha_\tau$ is the learning rate of the actor network.

The target networks\textquoteright \ weights might be updated by hard replacement or soft replacement procedures. The hard replacement copies the weights of the main networks to the target networks, while the soft replacement changes the weights of the target networks by a proportion $\lambda \in (0,1)$ of the main networks\textquoteright \ weights; that is,
\begin{equation}\label{eq:TargetCriticSoftReplacement} 
\theta ^{Q^\prime } \leftarrow \lambda^Q \theta ^{Q} + (1-\lambda^Q)\theta ^{Q^\prime }
\end{equation}
\begin{equation}\label{eq:TargetActorSoftReplacement}
\theta ^{\tau^\prime } \leftarrow \lambda^\tau \theta ^{\tau} + (1-\lambda^\tau)\theta ^{\tau^\prime }
\end{equation}
Random experience replay is regularly utilized with this algorithm by randomly sampling from the historical data stored in a memory to diminish biases from strongly correlated transitions and reduce training time. DDPG is summarized in Algorithm~\ref{Alg:DDPG}.

\begin{algorithm}
\SetKwInOut{Input}{Input}
\SetKwInOut{Output}{Initialization}
\Input{Maximum number of episodes ($\mathcal{E}$), Maximum number of time steps in one episode ($T$), replay memory $\mathbf{P}$, mini-batch size $\mathcal{M}$}
\Output{Randomly initialize networks' weights $\theta ^{\tau}$ and $\theta ^{Q}$, and target networks' weights $\theta ^{\tau^\prime} \gets \theta ^{\tau}$ and $\theta ^{Q^\prime} \gets \theta ^{Q}$ }
\For {$e = 1$ \KwTo $\mathcal{E}$}
{
    $t=1$\\
    Get initial state $s_1$.\\
    Initialize random noise $\mathbf{N}$.\\
    \While {$t \leq T$ and target state is not achieved}
    {
        Select action $a_t$ using actor network $\theta ^{\tau}$.\\
        Add noise: $a_t = a_t + \mathbf{N}$.  \\
        Execute $a_t$ and get next state $s_{t+1}$ and reward $r_t$.\\
        Store transition $(s_t,a_t,r_t,s_{t+1})$ in $\mathbf{P}$.\\
        \If {$t > \mathcal{M}$}
        {
            Randomly sample $\mathcal{M}$ transitions from $\mathbf{P}$.\\
            Train on batch of $\mathcal{M}$ transitions and compute target value according to equation~\ref{eq:ComputeTargetValue}.\\
            Update critic and actor networks according to equations~\ref{eq:UpdateCriticNetwork} and~\ref{eq:UpdateActorNetwork}.\\
            Update target critic and target actor networks according to equations~\ref{eq:TargetCriticSoftReplacement} and~\ref{eq:TargetActorSoftReplacement}.
        }
        $t \leftarrow t + 1$\\
    }
}
    
\caption{Deep Deterministic Policy Gradient (DDPG) Algorithm}\label{Alg:DDPG}
\end{algorithm}

\subsection{A Hierarchical Framework of DDPGs}\label{section-3.2}

\begin{figure}[ht]
\centering
\includegraphics[width=0.6\linewidth]{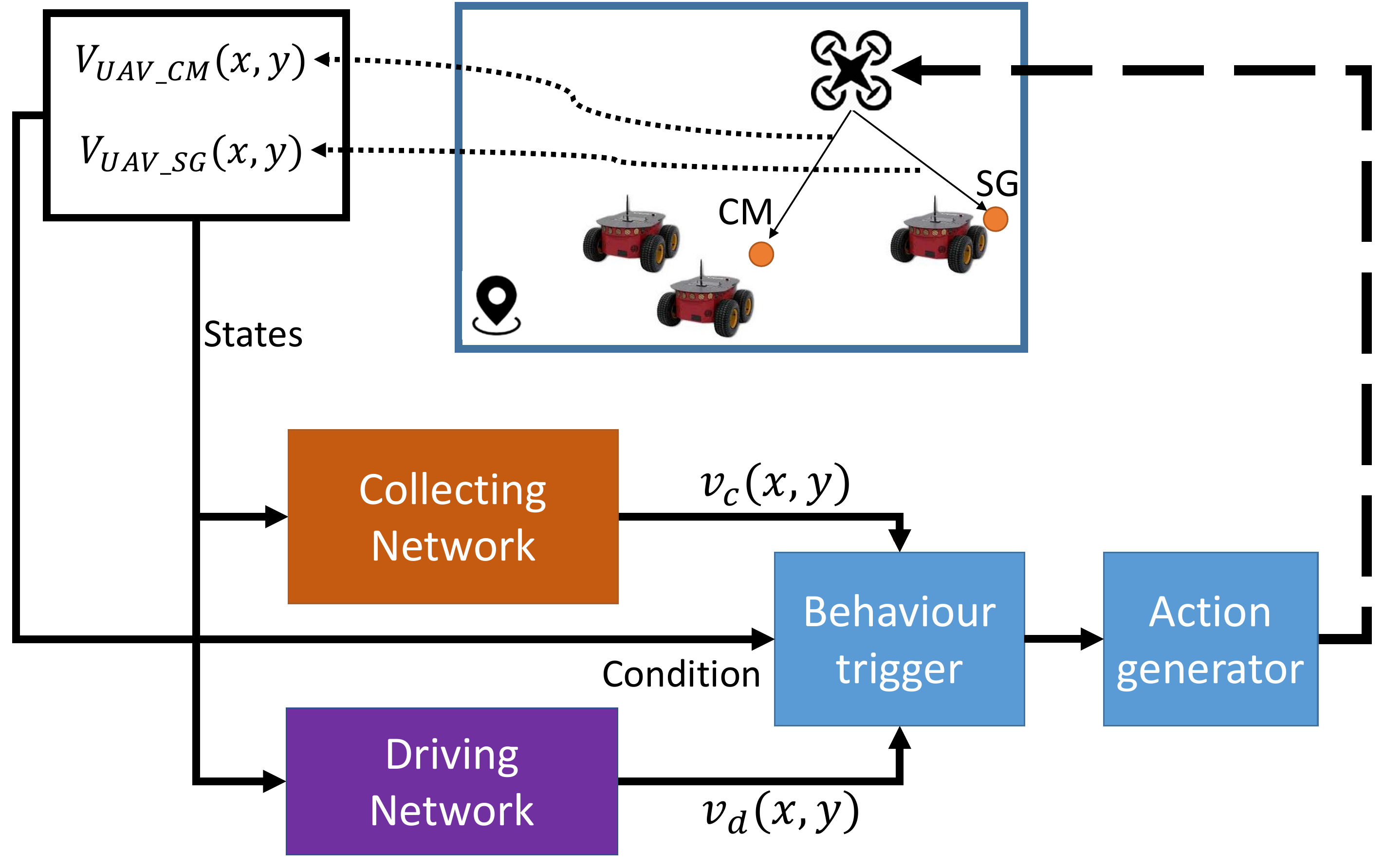}
\caption{Deep networks for controlling UAV in aerial shepherding scenario.}
\label{fig:Framework}
\end{figure}

In Figure~\ref{fig:Framework}, a hierarchical framework of deep RL using two DDPG networks for controlling the shepherding UAV is illustrated. The input of the networks are the relative directional vectors between the UAV and the sub-goal location (collecting/driving point), and between the UAV and the center of mass of UGVs. These inputs are calculated directly from sensor data and are used for training the deep neural networks.

\begin{figure}[ht]
\centering
\includegraphics[width=0.5\linewidth]{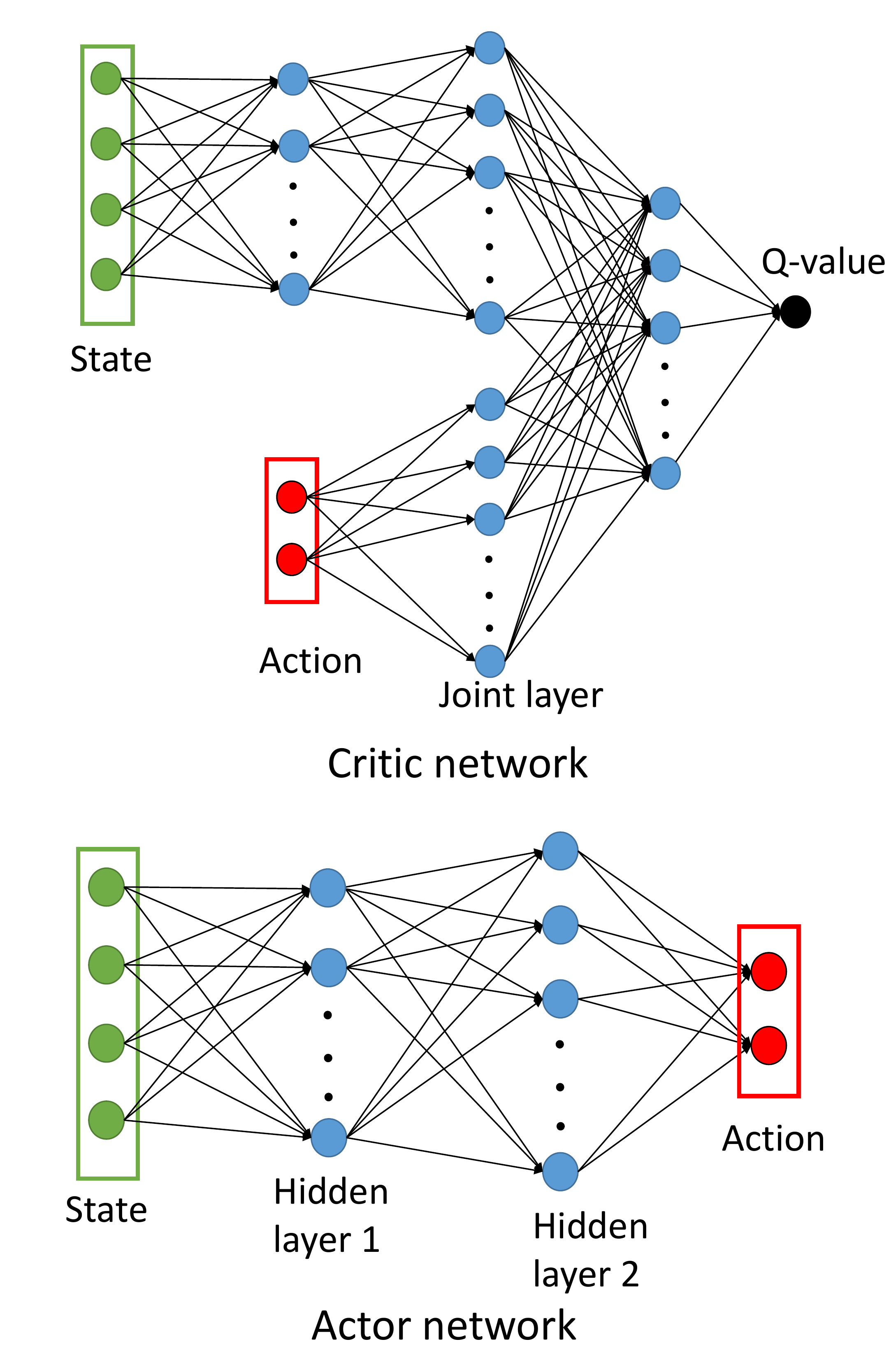}
\caption{DDPG networks architecture.}
\label{fig:DDPG}
\end{figure} 

\subsubsection{Skill Decomposition Framework}\label{section-3.2.1}

Given the complexity of the problem, it is challenging for an agent to learn to control a swarm of other agents to complete multiple mission\textquoteright s requirements~\cite{Gebhardt2018learning}. Thus, using a single deep reinforcement learning network to achieve optimal behavior for different objectives is difficult due to the large search space. Our previous work decomposed the shepherding learning problem to two sub-problems: one for collecting and the other for driving. The UAV can learn to complete a shepherding mission by learning these two sub-problems independently; thus, the complexity of the learning problem becomes more manageable~\cite{nguyen2018swarm,nguyen2018apprenticeship}. 

Two independent training sessions are conducted simultaneously, one for learning to collect and the second for learning to drive. In the former, the training session begins with the initialization of an environment where a UGV is situated at a position far away from a cluster of the other UGVs. For every time step, the UAV learns to navigate to a collection point behind the furthest UGV from the group\textquoteright s center of mass. When the UAV reaches the collection point, the training session for collecting ends. The training session for driving is similar except that the UAV learns to reach a position on the vector from the goal to the UGVs center of mass, and outside the perimeter of the UGVs\textquoteright \ cluster.

During testing, the two trained networks get connected to a logical gate that switches between the two behaviors based on the state of the UGVs in the environment.

\subsubsection{Reward Design}\label{section-3.2.2}

Let $p_{UAV}^t = <x_{UAV}^t,y_{UAV}^t>$ and $p_{subgoal}^t = <x_{subgoal}^t,y_{subgoal}^t>$  be the position of the UAV and the subgoal at time $t$. The distance between them is computed by $d^t_{sg} = \|p_{subgoal}^t - p_{UAV}^t\|$. When the UAV moves with a velocity of $v^t_{sg}(x,y)$, the next position of the UAV is $p_{UAV}^{t+\delta t} = <x_{UAV}^t + v^t_sg(x)*\delta t,y_{UAV}^t + v^t_{sg}(x)*\delta t >$. The next distance of the UAV to the subgoal will be $d^{t+\delta t}_{sg} = \|p_{subgoal}^{ t+\delta t} - p_{UAV}^{ t+\delta t }\|$.  A positive reward is received if $ (d^{t+\delta t}_{sg} - d^{t}_{sg}) \leq 0$, i.e. the UAV gets closer to the sub-goal, and a negative reward is received otherwise. The reward is discounted over time, therefore the UAV has to learn to optimize its course of actions to navigate to the sub-goal as fast as possible.
\begin{equation}
    W^t = \begin{cases} 0.1, & \mbox{for } d^{t+\delta t}_{sg} - d^{t}_{sg} \leq 0 \\ -0.1, & \mbox{otherwise} \end{cases}
\end{equation}
In the testing scenario, a large reward is received when the whole mission is completed.

\section{Experimental Setups}\label{section-4}

In this paper, we test our proposed model in both simulation and physical environments. We investigate the learning success rate and the training environment\textquoteright s scalability, where we evaluate how the algorithm performs when the size of the training environment during simulation is different from the physical environment. While, in hindsight, this sounds like a simple rescaling that needs to occur, the scaling problem is non-trivial due to the coupling between the scale and the non-linear functional approximation of DDPG.

\subsection{Experimental Design}\label{section-4.1}

In our shepherding scenario, there is one UAV acting as a sheepdog and three UGVs acting similar to sheep. An underlying assumption is that the UAV has global-sensing ability which enables it to sense the positions of all the UGVs in the environment. In practice, this means that either the UGVs are within the camera field of view of the UAV or they are within the communication range of the UAV. The UAV is required to drive the swarm of the UGVs to a target. 

\subsubsection{Action and state space}\label{section-4.1.1}
For the UGVs, the action space consists of two continuous values representing the linear velocity, $V_{UGVs}$, and the angular velocity (yaw rate), $\omega$. This velocity represents the step length of a sheep per second. Similarly, for the UAV, the action space consists of two real values representing the linear velocity, $V_{UAV}$, according to the longitudinal and lateral directions. The commanded speed of the UAV varies in the range $[-1,1]$ $m/s$. At each episode in both the training and testing processes, the UAV automatically takes off to a height of $2m$, and the height is maintained till the end of the episode. While the height of a UAV would change the influence zone on the ground vehicle, it does not impact the ability of the AR.Drone quadrotor to track a complicated reference trajectory along the $x$ and $y$ axes as demonstrated in \cite{santana2015outdoor,santana2014trajectory}.
The state inputs include $x$ and $y$ coordinates of two representational vectors: UGVs\textquoteright \ centre of mass to UAV ($GCMtoD(x,y)$) and Subgoal to UAV ($SubtoD(x,y)$)

\subsubsection{Deep neural network structure}\label{section-4.1.2}

In the paper, we use a deep feedforward architecture for DDPG. The deep actor network includes two hidden layers with 32 and 64 nodes while the deep critic network is more complex as shown in Figure~\ref{fig:DDPG}. The optimizer method is Adam~\cite{kingma2014adam}, and the training parameters are the same used in the original DDPG algorithm~\cite{mnih2015human}. The H-DDPG algorithm is trained with a replay memory size of $10^5$ state-action pairs, the discount factor $\gamma = 0.99$, mini-batch size 32, and the learning rate for the actor network and the critic network being 0.00025 and 0.001, respectively. The maximum number of steps in both collecting and driving DDPG models is 1000.

\subsubsection{Experimental setups}\label{section-4.1.3}
In the simulation experiment, we aim to investigate the learning success rate and the scalability of the training environment when training our proposed H-DDPG approach in both 4$\times$4 and 6$\times$6 environments. Firstly, we train the collecting and driving DDPG models using both environment sizes. Then, we test the models trained in the 4$\times$4 environment on the 6$\times$6 environment. The learning framework can demonstrate the scalability when an agent, which is successfully trained in a small simulation environment, can be effectively applied in a larger simulation or physical environment. When adopting the model trained from the smaller environment into a larger environment, the inputs and outputs of the deep networks are scaled. The scaling factor is calculated as in Equation~\ref{eq:scale}.
\begin{equation}\label{eq:scale}
\xi = \frac{\sqrt{(E^{small}_x)^2 + (E^{small}_y)^2}}{\sqrt{(E^{big}_x)^2 + (E^{big}_y)^2}}
\end{equation}
where $E^{small}$ and $E^{big}$ represent the sizes of small and large environments. $E_x$ and $E_y$ are width and height of the environment. The state input and action output of the model are multiplied with $\xi$ and $1/\xi$, respectively. The action output was normalized using \textit{tanh} to limit the range between -1 to 1.

In this paper, we conduct two experimental setups. The first setup aims to evaluate the learning success rate and the scalability in response to changes in the size of the environment. The second setup tests the models trained in simulation in a physical environment. Table~\ref{tab:experimental-setups} shows the parameters used in both the setups. 

\begin{table}[ht]
\small
\caption {Parameters in setups.} \label{tab:experimental-setups}
 \begin{center}
  \begin{tabular}{lll}
    \hline
    \textbf{Name} &  \textbf{Parameters} \\\hline 
    $W_{\pi\pi}$  & 0.5 (m)\\
    $R_{\pi\pi}$  & 1 (m) \\
    $f(N)$  & 1.3 (m) \\
    $R_{\pi\beta}$  & 2 (m)\\
    $\mathbb{D}$  & 2 (m)\\
    $V_{UGVs}$ & 0.5 (m/s)\\
    $Time-Step$ & 0.1 (s)\\
    \hline
  \end{tabular}
 \end{center}
\end{table}

\subsubsection{Evaluation Metrics}\label{section-4.1.4}

Four metrics for evaluating the performance of the learning framework for the UAV-UGV shepherding problem are listed below.

\begin{itemize}
    \item \textbf{Cumulative reward} over training episodes that UAV agent receives.
    \item \textbf{Success rate ($\%$)} is the percentage of mission completion computed on 30 trials. The mission success is achieved when all the UGV are collected and driven to the goal position.
    \item \textbf{Number of steps} is the number of steps for the centre of mass of UGVs to reach goal position.
    \item \textbf{Travel Distance (m)} is the total distance that the UAV covers in the environment.
    \item \textbf{Error per step (m)} is the difference between the desired and actual positions meanwhile the desired position is the sum of the previous position with the desired movement. The desired movement is calculated by  multiplying the linear velocity with the time of a step.
    \item \textbf{Distance from the aerial shepherd and the sub-goal per step (m)}.
    \item \textbf{Reduced distance from the center of the mass of the UGVs and the target per step (m)}.
\end{itemize}

While the cumulative reward and success rate reflect the training and testing effectiveness of the learning algorithm, the remaining metrics represent the efficiency of the policy proposed by the learning framework. In addition, the trajectories of the UAV and UGVs\textquoteright \ center of mass in the simulation and physical environments are visualized to illustrate mission performance.

\subsection{Environment and Control Network Setups}\label{section-4.2}

Our proposed hierarchical framework using DDPG networks to control shepherding UAV is initially trained and tested within a simulation environment. Robot operating system (ROS) is used as an interface which allows the agents to communicate with the Gazebo simulation environment. The simulator package for the UAV is Tum-Simulator~\cite{huang2014tum}, which simulates the Parrot AR Drone 2.

We further evaluate the transferability of our proposed model to a physical environment. The control network in the physical environment includes a VICON motion capture system (MCS) to detect the states of the environment, a base station where the information is processed and command messages are automatically generated, and the UAV and UGVs. Detailed description and specifications of each component in the system can be found in Section~\ref{section-S1} (Supplementary Document).

\begin{figure}[ht]
    \begin{center}
        \includegraphics[width=0.6\linewidth]{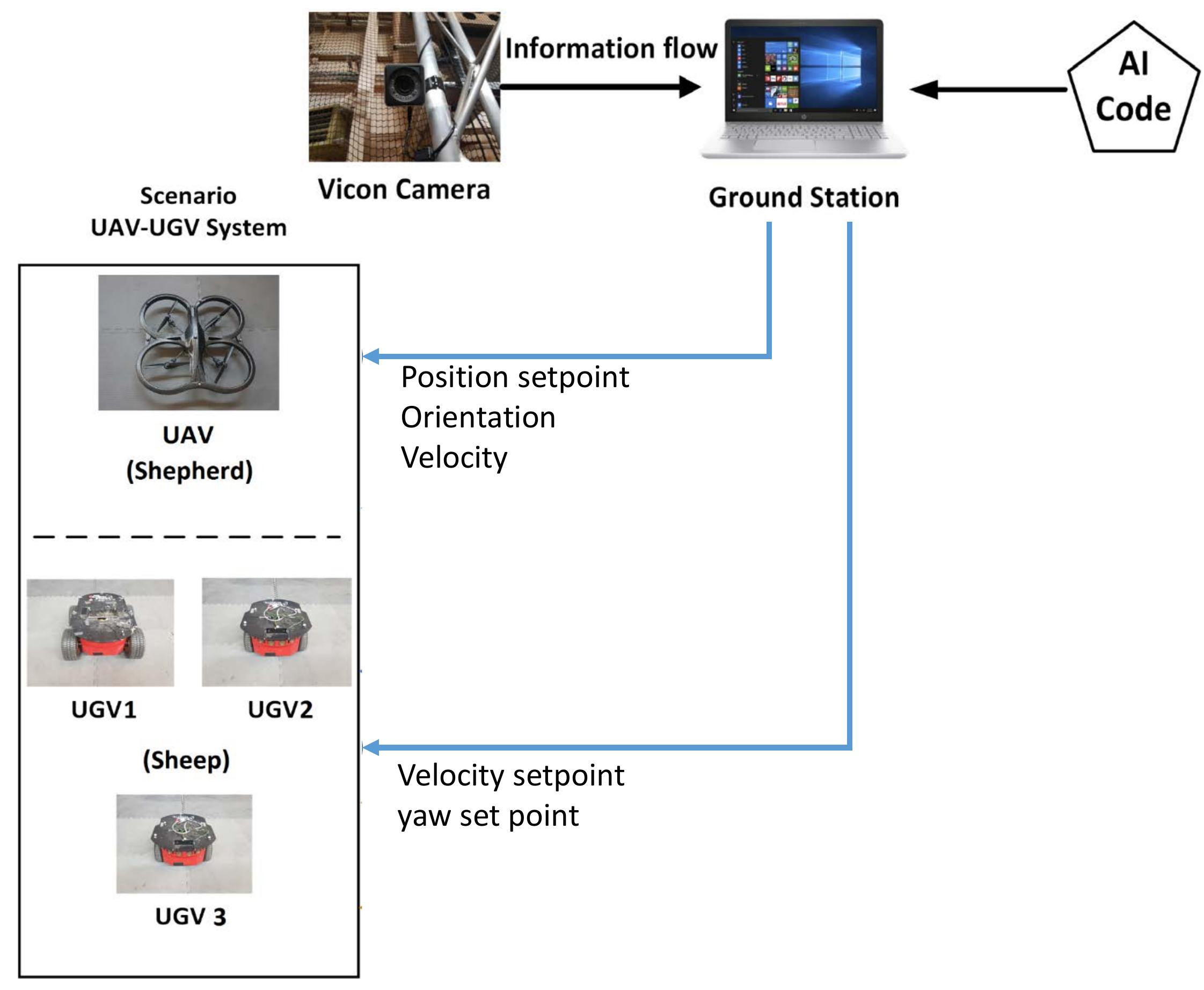}    
        \caption{\footnotesize Overall network architecture.}  
        \label{fig:network_architecture}                                 
    \end{center}                                 
\end{figure}

Figure~\ref{fig:network_architecture} shows how data is transferred in the physical system. Firstly, the VICON MCS broadcasts continuously to each entity at a frequency of 100 Hz using a UDP network protocol. Regarding the control network, the central computer, containing the AI code, receives all UAV and UGVs\textquoteright \ states being the position and orientation in order to calculate the input states for the AI program. After that, the AI program produces commands [$velx, vely, \tilde{\psi}$] for the UGVs and [$\dot{\phi}, \dot{\theta}$] for the UAV using ROS messages.

\section{Evaluation in Simulation}\label{section-5}

We compare the performance of the H-DDPG with that of the DHRL and the Str\"{o}mbom approach as a baseline method. For the DHRL~\cite{nguyen2019deep}, we re-trained the two deep Q-network (DQN) models (driving and collecting) until convergence with the same parameters of the shepherding task shown in~\ref{tab:experimental-setups}. In the testing phase, our proposed H-DDPG are tested on 30 different testing cases. In each testing case, the UAV is initialized at a different position in the simulation environment. Both the DHRL and Str\"{o}mbom methods are tested in this testing set.

\subsection{Training}\label{section-5.1.1}
\begin{figure*}[!ht]
    \centering
    \subfigure[Driving in 4$\times$4]
    {
        \includegraphics[width=0.4\textwidth]{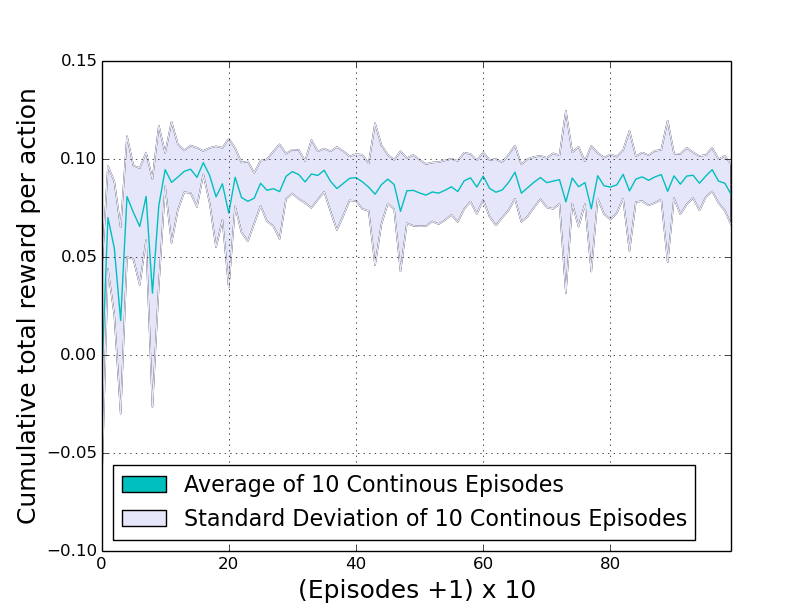}
        \label{fig:driving4x4_sub}
    }%
    \subfigure[Collecting in 4$\times$4]
    {
        \includegraphics[width=0.4\textwidth]{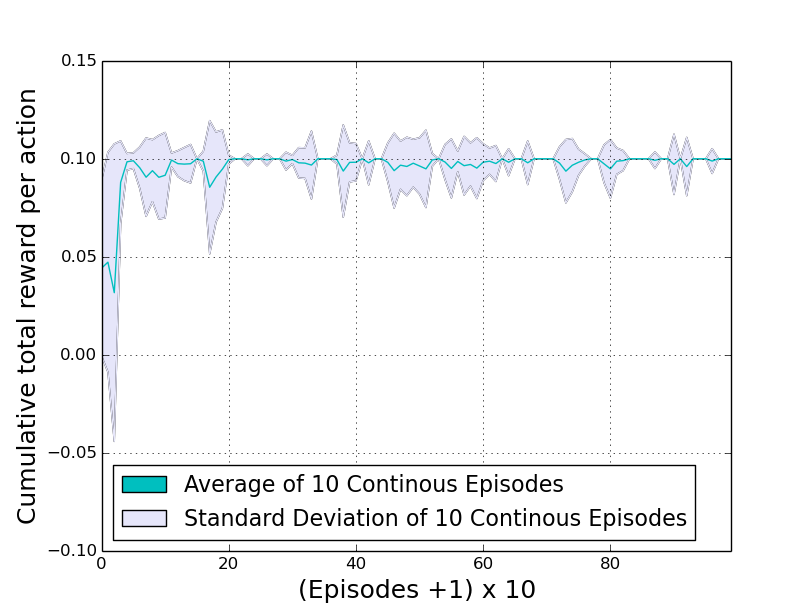}
        \label{fig:collecting4x4_sub}
    }%
    \\
    \subfigure[Driving in 6$\times$6]
    {
        \includegraphics[width=0.4\textwidth]{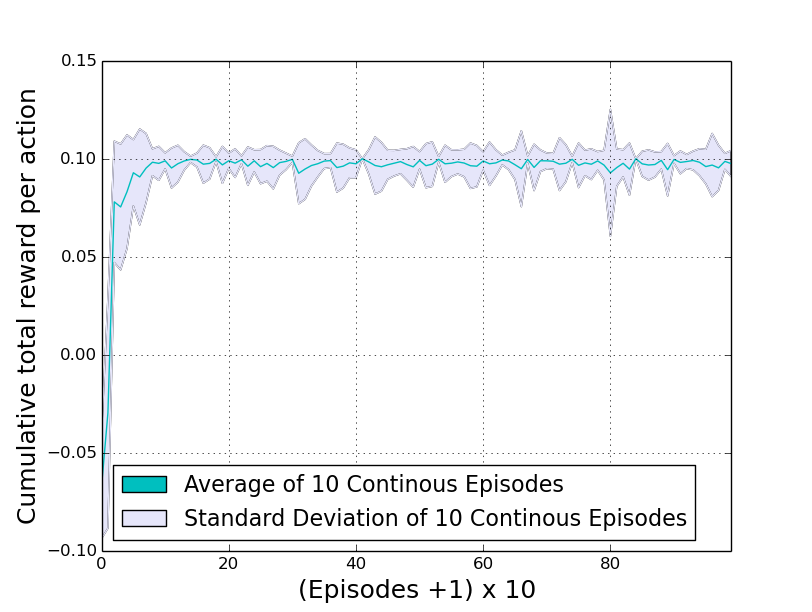}
        \label{fig:driving6x6_sub}
    }%
    \subfigure[Collecting in 6$\times$6]
    {
        \includegraphics[width=0.4\textwidth]{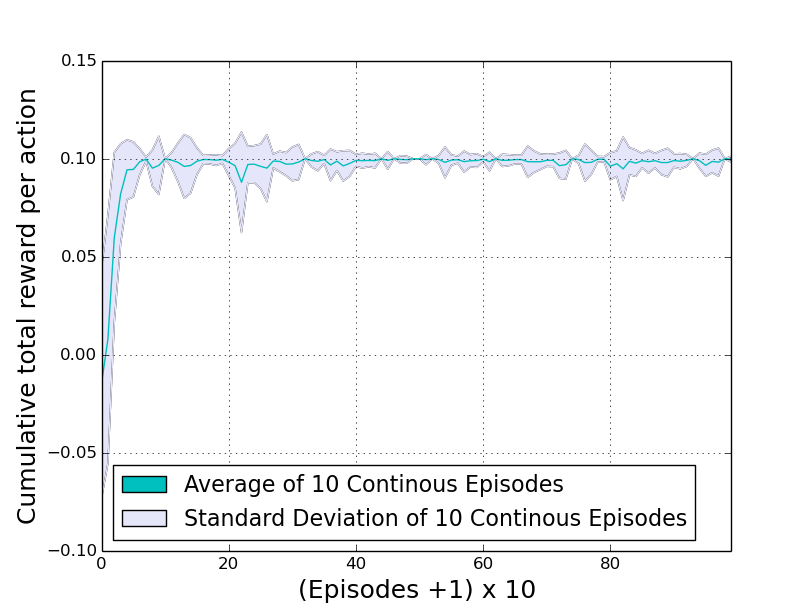}
        \label{fig:collecting6x6_sub}
    }
    \caption{Learning curves of two driving and collecting DDPG models on 4$\times$4 and 6$\times$6 environments.}
    \label{fig:learningcurves}
\end{figure*}

The two collecting and driving DDPG models of our H-DDPG approach in both 4$\times$4 and 6$\times$6 environments are fully trained in 3000 episodes. Figure~\ref{fig:learningcurves} show averages and standard deviations of cumulative total reward per action in every 10 episodes.

The cumulative total rewards per action of both the collecting and driving models increase significantly in the first 100 episodes, and then becomes relatively stable with a value of approximately 0.09 till the end of training. The tendencies demonstrate that these collecting and driving DDPG models are able to learn effectively when they converge at an approximate reward of 0.085.
This tendency shows the convergence of the learning processes of four DDPG models in both the 4$\times$4 and 6$\times$6 environments.

\subsection{Testing}\label{section-5.2}
The testing shepherding environment is the same as the training environment. In total, we conduct eight testing scenarios as shown in Table~\ref{tab:simulationexperiments}. For the transition in testing scenario between DHRL-4$\times$4 to 6$\times$6, we only scale the state because of the fixed discrete action output of the DHRL in both the two environment. 
\begin{table}[ht]
\small
\caption {Testing scenarios on 4$\times$4 and 6$\times$6 environments.} \label{tab:simulationexperiments}
 \begin{center}
  \begin{tabular}{ll}
    \hline
    \textbf{Testing ID} &  \textbf{Description } \\\hline 
    Strombom-4$\times$4& Testing the Strombom model on 4$\times$4\\
    Strombom-6$\times$6 & Testing the Strombom model on 6$\times$6\\
    DHRL-4$\times$4& Testing the 4$\times$4 trained model on 4$\times$4\\
    DHRL-6$\times$6 & Testing the 6$\times$6 trained model on 6$\times$6\\
    DHRL-4$\times$4to6$\times$6& Testing the 4$\times$4 trained model on 6$\times$6 with scale.\\
    H-DDPG-4$\times$4& Testing the 4$\times$4 trained model on 4$\times$4\\
    H-DDPG-6$\times$6 & Testing the 6$\times$6 trained model on 6$\times$6\\
    H-DDPG-4$\times$4to6$\times$6& Testing the 4$\times$4 trained model on 6$\times$6 with scale.\\
    \hline
  \end{tabular}
 \end{center}
\end{table}

In each testing scenario, we conduct the 30 different testing cases, and then calculate average and standard deviations of the number of steps, the travel distance, and success rate. 

Table~\ref{tab:experrors-sim4x4} shows the results of the three methods tested in the 4$\times$4 environment. H-DDPG outperforms the Str\"{o}mbom approach on the three metrics (the number of steps, traveled distances, and success rate). However, in the larger environment of 6$\times$6, it seems that the Str\"{o}mbom model performs better than the learning methods in both the number of steps and the travelled distance, but not in terms of success rate.

\begin{figure*}[!ht]
    \centering
    \subfigure[Strombom-6$\times$6]
    {
        \includegraphics[width=0.4\linewidth]{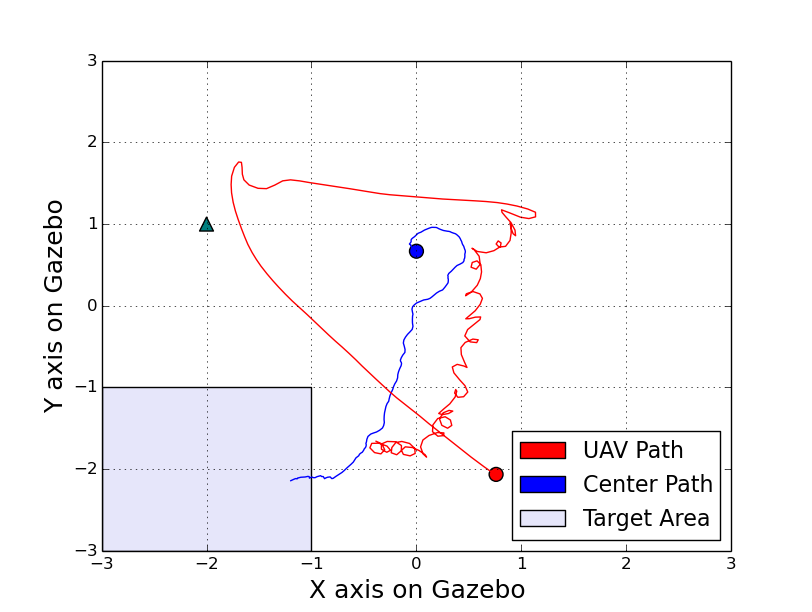}
        \label{fig:traj-sim-6x6-Stom}
    }%
    \subfigure[DHRL-6$\times$6]
    {
        \includegraphics[width=0.4\linewidth]{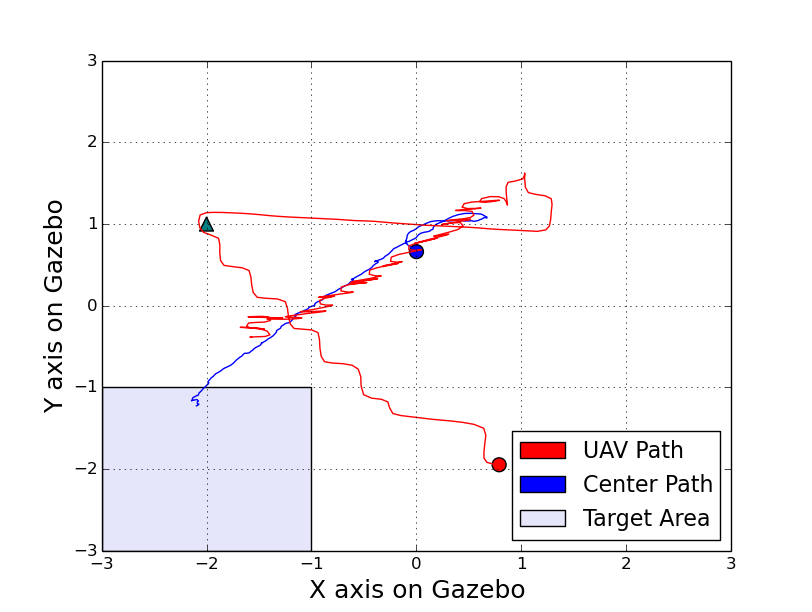}
        \label{fig:traj-sim-6x6-DHRL}
    }%
    \\
    \subfigure[HDDPG-6$\times$6]
    {
        \includegraphics[width=0.4\linewidth]{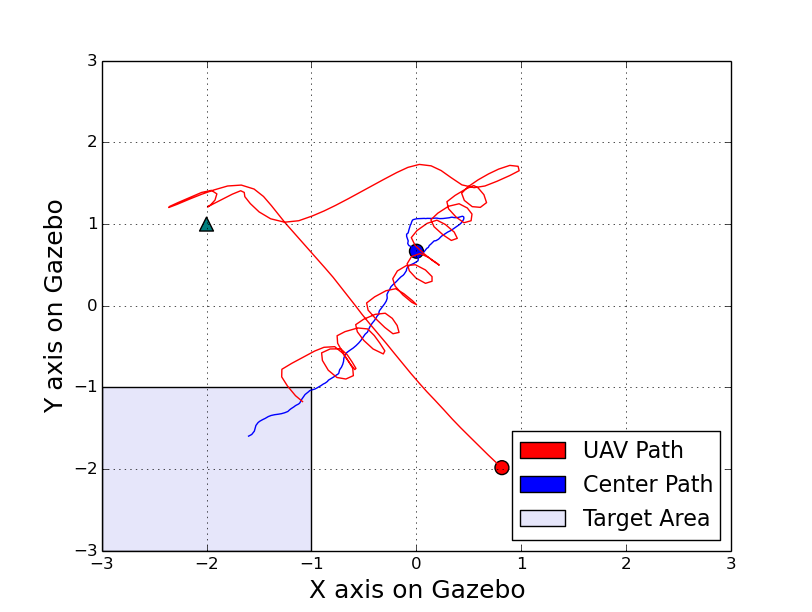}
        \label{traj-sim-6x6-HDDPG}
    }%
    \subfigure[HDDPG-4$\times$4to6$\times$6]
    {
        \includegraphics[width=0.4\linewidth]{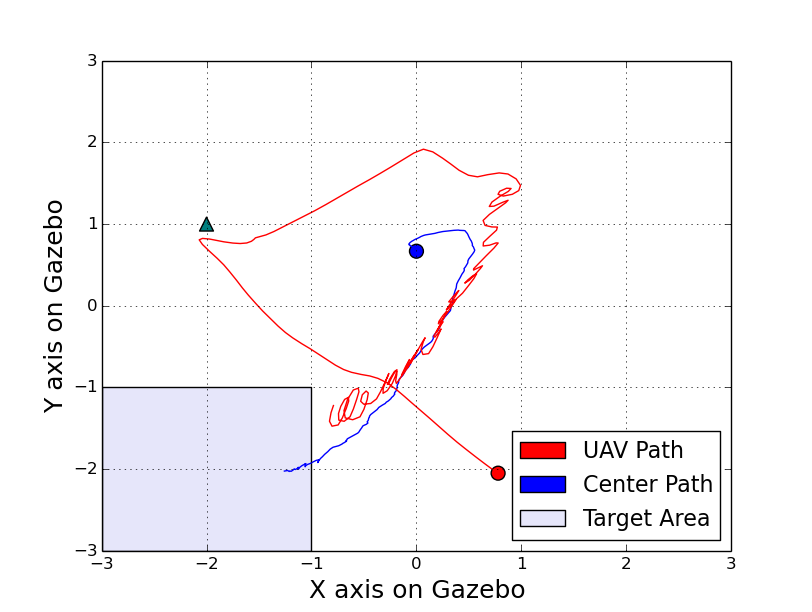}
        \label{traj-sim-4x4to6x6-HDDPG}
    }
    \caption{Trajectories of the UAV and the center of mass in the 15th testing case performed in a 6$\times$6 simulation environment. The movement of DHRL is zigzag being less smoother that of HDDPG. The movement of HDDPG in the 4$\times$4to6$\times$6 environment is narrower than that of the HDDPG trained in the 6$\times$6 environment. Compared to the Str\"{o}mbom model, the behaviour of the three trained models seems to drive better when the mass of the UGVs are guided straight to the target.  }\label{fig:trajectories_sim_4x4and6x6}
\end{figure*}

H-DDPG used slightly less number of steps than DHRL in both the 4$\times$4 and 6$\times$6 environments. Although the travelled distance of the H-DDPG is slightly longer, the discrete action space of DHRL is unrealistic and therefore, the shorter distance reflects non-smooth and sharper manoeuvres by the UAV. The discrete actions of the DHRL produces unnatural zigzag movements reducing the travelled distance. The difference in the two movements is shown in~Figure~\ref{fig:trajectories_sim_4x4and6x6}.

Table~\ref{tab:experrors-sim6x6} shows that after scaling both actions and states, the trained 4$\times$4 H-DDPG model outperforms the H-DDPG model trained in the larger environment of 6$\times$6 in terms of the travelled distance, but the first model takes some more steps. It can be understood that the action values of 4$\times$4 H-DDPG model is slightly smaller than that of the 6$\times$6 model when these values are scaled and then put through the function \textit{tanh}. These smaller values helps to reduce the wide movement of the UAV, causing the path of the 4$\times$4 H-DDPG agent to be slightly shorter. This view is illustrated in Figure~\ref{fig:trajectories_sim_4x4and6x6}.
 
From these testing results in the simulation, it is clear that the H-DDPG learning approach produces agents which are able to successfully aerial shepherd. Additionally, the results show that it is feasible that an agent trained in a small environment has the ability to perform in a larger environment with the same task or even for the aerial shepherding task.

\begin{table}[!ht]
\small
 \centering
\caption {Averages and standard deviations of number of steps, travelled distance, and success rates of the three setups in 30 testing cases in 4x4 simulation environment.} 
\label{tab:experrors-sim4x4}
  \centering
  \begin{tabular}{llll}
    \hline
     \textbf{Experiments} &  \textbf{Number-Steps} & \textbf{Travelled-Distance} &  \textbf{Success}\\
      & MSE $\mu \pm \sigma$ & MSE $\mu \pm \sigma$ &(\%)\\  
    \hline
Strombom-4$\times$4 &  135    $\pm$  119   &  11    $\pm$  9.3 &  96.67 \\
DHRL-4$\times$4 &  107    $\pm$  16   & 8.3    $\pm$   1.2 & 100 \\
H-DDPG-4$\times$4&  \textbf{100    $\pm$  16}   &9.7    $\pm$   1.5 & 100\\
\hline
  \end{tabular}
\end{table}

\begin{table}[!ht]
\small
 \centering
\caption {Averages and standard deviations of number of steps, travelled distance, and success rates of the five setups in 30 testing cases in 6x6 simulation environment.} 
\label{tab:experrors-sim6x6}
  \centering
  \begin{tabular}{llll}
    \hline
     \textbf{Experiments} &  \textbf{Number-Steps} & \textbf{Travelled-Distance} &  \textbf{Success}\\
      & MSE $\mu \pm \sigma$ & MSE $\mu \pm \sigma$ &(\%)\\  
    \hline
Strombom-6$\times$6 &  187    $\pm$  21   &  15.2    $\pm$  1.8 &  96.67 \\
DHRL-6$\times$6 &  224    $\pm$  32   & 16.9    $\pm$   2.4 & 100 \\
DHRL-4$\times$4to6$\times$6 &  206    $\pm$  29   & 15.9    $\pm$   2.3 & 100 \\
H-DDPG-6$\times$6&  \textbf{188    $\pm$  44}   &19.9    $\pm$   5.4 & 100\\
H-DDPG-4$\times$4to6$\times$6 &  205    $\pm$  20   & \textbf{16.7    $\pm$   1.8} & 100 \\
\hline
  \end{tabular}
\end{table}

The promising results in the simulation initially demonstrate the effectiveness of our proposed learning approach for the aerial shepherding task. Besides, they also show that it is not necessary to train the agent in the same simulation environment and that the skills learnt, with appropriate re-scaling, are usable. When the agent was trained in the smaller environment, it still had an ability to perform well or even better.

\section{Evaluating on physical system}\label{section-6}
Transferring the learnt models from simulation to the physical environment is challenging. Firstly, the yaw of the UGVs needs to be stabilized in order to make them move in the required directions without violating the non-holonomic constraints for a wheeled vehicle. 

Secondly, our proposed algorithm gives virtual force commands which basically correlate to an acceleration command. We effectively integrate these to get velocity and again to get the position we use the prescribed position command. The UAV quadcopter is controlled to follow a desired position trajectory rather than following the acceleration references in order to reduce the tracking errors, caused by offset errors and exogenous disturbances. In the next section, we describe these solutions in the physical environment before conducting the testing scenarios for our learning approach.

\subsection{System stabilization}\label{section-6.1}
\subsubsection{UGVs yaw stabilization}\label{section-6.1.1}
The UGV absolute yaw angle was controlled by an adaptive Strictly Negative Imaginary (SNI) controller-based Fuzzy Interference System (FIS) whose parameters were properly tuned by minimizing the tracking errors through trial-and-error~\cite{tran2019}. To do so, we first rotated the yaw axis in VMTS to coincide with the one in the Gazebo simulation. After that, we designed the closed-loop yaw control loop to stabilize the rotational motion of all mobile robots. Descriptions of the Fuzzy-SNI control strategy and a test experiment for control stabilization can be found in Section~\ref{section-S2} (Supplementary Document).

\subsubsection{UAV control stabilization}\label{section-6.1.2}
Since the AR. Drone quadcopter is subject to multiple disturbances and has control offsets which can cause position drift, controlling the AR. Drone quadcopter based on velocity commands is not feasible in areas with restricted space~\cite{fernandez2017}, especially in the $6m\times6m$ area found in our VICON lab.

To solve this problem, we estimated the next global position based on the UAV\textquoteright s current coordinate and its velocity setpoints as in Equation~\ref{eq:41} and then stabilized the UAV position using the adaptive Strictly-Negative Imaginary (SNI) position tracking controller as described in~\cite{tran2017, tran2019,tran2020switching}. As soon as a desired position in waypoints was reached, the quadcopter starts to receive the next velocity references and produce the next desired position using Eq. (\ref{eq:41}).
\begin{equation}\label{eq:41}
\tilde{p}_{x,y}(k+1) = \tilde{p}_{x,y}(k)+V_{x,y}*dt,
\end{equation}
where ($\tilde{p}_{x,y}$ denotes the desired position of the UAV on the planar plane. Next, $V_{x,y}$ highlights the actual velocity of the UAV along the $x$ and $y$ axis. While $k$ is the time step, $dt$ is the sample time.

\subsection{Testing}\label{section-6.2}
We adopt the parameters as shown in Table~\ref{tab:experimental-setups} except for the time step. For this setup, we set the time step at $0.2$ seconds in both the simulation and physical environment. Firstly, we conduct testing scenarios in the simulation, and then in the physical environment with the same parameters.
Four testing scenarios are shown in Table~\ref{tab:physicalexperiments}.
\begin{table}[ht]
\small
\caption {Testing scenarios on 6$\times$6 simulation and physical environments.} \label{tab:physicalexperiments}
 \begin{center}
  \begin{tabular}{ll}
    \hline
    \textbf{Testing ID} &  \textbf{Description } \\\hline 
    HDDPG-6$\times$6-Sim & Testing the 6$\times$6 trained model\\
    &on 6$\times$6 simulation\\
    HDDPG-4$\times$4to6$\times$6-Sim& Testing the 4$\times$4 trained model\\
    &on 6$\times$6 simulation with scale.\\
    HDDPG-6$\times$6-Phy & Testing the 6$\times$6 trained model \\
    &on 6$\times$6 physical\\
    HDDPG-4$\times$4to6$\times$6-Phy& Testing the 4$\times$4 trained model\\
    &on 6$\times$6 physical with scale.\\
    \hline
  \end{tabular}
 \end{center}
\end{table}

We calculate average and standard deviations of the number of steps, the travel distance, the error per step between the desired movement and the actual movement, the distance per step between the position of the aerial shepherd and the sub-goal, the reduced distance per step between the center of the mass of the UGVs and the target, and the success rate as shown in Table~\ref{tab:experrors-phys}. 

The error per step of each episode is calculated by measuring the distance between the actual position and the desired position in every step. 
For both the simulations and real experiments, the UAV is stabilized by a position tracking system. The desired position is updated by adding the desired linear velocity multiplied by the time step ($0.2$ second) at each time step.  


Table~\ref{tab:experrors-phys} show interesting results in the testing scenarios. Firstly, the success rate of the four testing scenarios is $100\%$ when all the agents pass three different testing cases in both simulation and physical experiments. Similar to the investigation of the previous testing setup, Table~\ref{tab:experrors-phys} shows that the 4$\times$4 agent after being scaled by Equation~\ref{eq:scale} is able to perform better than the agent trained in the 6$\times$6 agent, and in the 6$\times$6 testing environment. In the simulation, the travel distance of the 4$\times$4 agent is $13.4m$ compared to $17.4m$ of the 6$\times$6 agent. When we inspect the error per step of the two agents, the error of the 4$\times$4 agent is considerably smaller than that of the 6$\times$6 agent, which are $0.018m$ and $0.026m$, respectively. Although the distance per step is larger, it is not enough to impact the entire performance. This higher distance per step appears because of the different size of the environment even though the agent\textquoteright s actions are scaled up. The reduced distance per step between the center and the target of the testing scenarios in the simulation is the same $(0.006m)$.

\begin{table}[!ht]
 \centering
\caption {Averages and standard deviations of number of steps, travelled distance, cumulative error and success rates of three testing cases of the four setups in simulation and physical environments.(NS- Number of Steps; TD - Travelled Distance; EPS - Error-Per-Step; DS - Dog-to-Subgoal; CT - Center-to-Target; SR - Success Rate)} 
\label{tab:experrors-phys}
  \centering
  \tiny
  \begin{tabular}{lllllll}
    \hline
     \textbf{Experiments} &  \textbf{NS} & \textbf{TD} & \textbf{EPS} &  \textbf{DS}&\textbf{CT} &\textbf{SR}\\
      & MSE $\mu \pm \sigma$ & MSE $\mu \pm \sigma$ & MSE $\mu \pm \sigma$ & MSE $\mu \pm \sigma$ & MSE $\mu \pm \sigma$ &(\%)\\  
    \hline
HDDPG-6$\times$6-Sim &  505    $\pm$  23   &  17.4    $\pm$   1.4& 0.026    $\pm$  0.003 & 0.63    $\pm$  0.03 &  0.006    $\pm$  0.000 &100 \\
HDDPG-4$\times$4to6$\times$6-Sim &  533    $\pm$  55   &13.4   $\pm$   0.9 & 0.018 $\pm$0.001 & 0.72    $\pm$  0.04 &  0.006    $\pm$  0.000 & 100\\
HDDPG-6$\times$6-Phy &  911    $\pm$  93   &  27.6    $\pm$   2.6 & 0.05    $\pm$  0.000 & 1.35    $\pm$  0.05 &  0.003    $\pm$  0.000 & 100\\ 
HDDPG-4$\times$4to6$\times$6-Phy &  881    $\pm$  88   &24.1    $\pm$   1.7 & 0.05 $\pm$0.001& 1.29    $\pm$  0.06 &  0.003    $\pm$  0.000 & 100\\ 
\hline
  \end{tabular}
\end{table}

\begin{figure}[!ht]
    \centering
    \subfigure[HDDPG-6x6to6x6-Sim:Trajectory]
    {
        \includegraphics[width=0.25\textwidth]{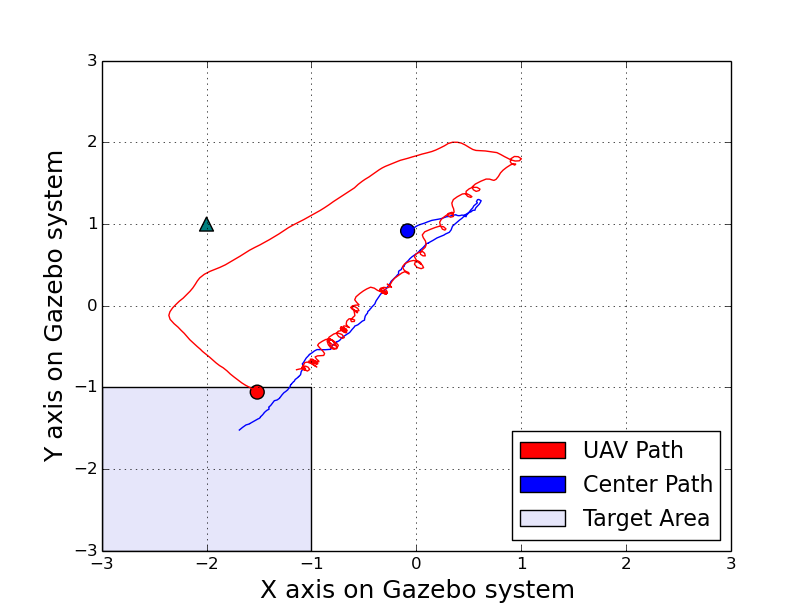}
        \label{fig:6x6to6x6-sim-traj}
    }%
    \subfigure[HDDPG-4x4to6x6-Sim:Trajectory]
    {
        \includegraphics[width=0.25\textwidth]{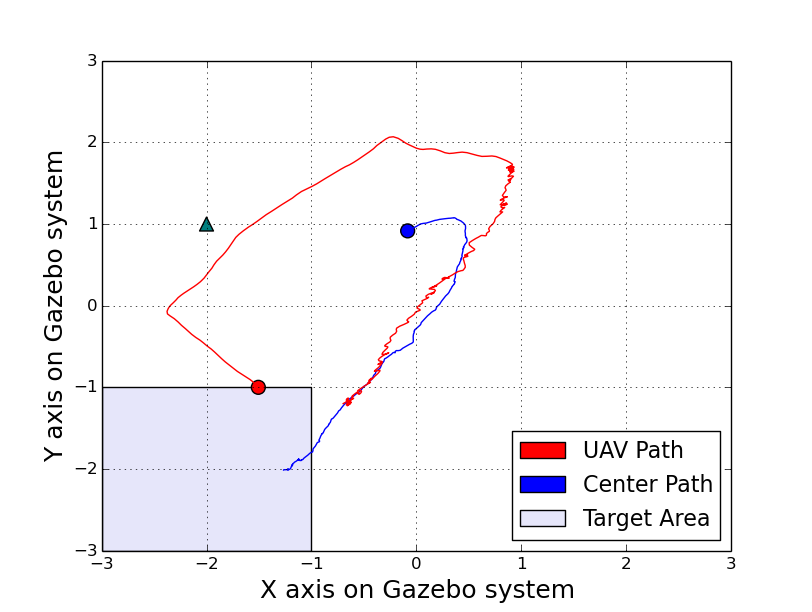}
        \label{fig:4x4to6x6-sim-traj}
    }%
     \subfigure[HDDPG-6x6to6x6-Phy:Trajectory]
    {
        \includegraphics[width=0.25\textwidth]{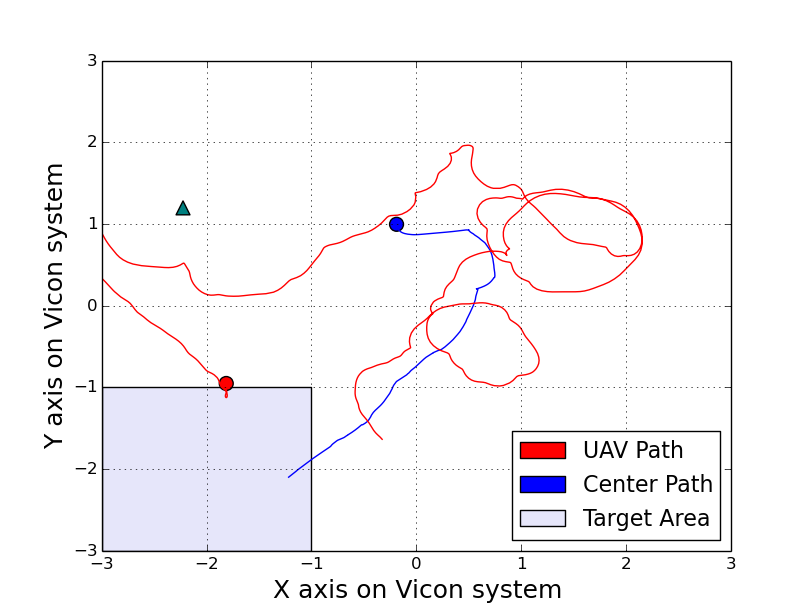}
        \label{fig:6x6to6x6-phy-traj}
    }%
    \subfigure[HDDPG-4x4to6x6-Phy:Trajectory]
    {
        \includegraphics[width=0.25\textwidth]{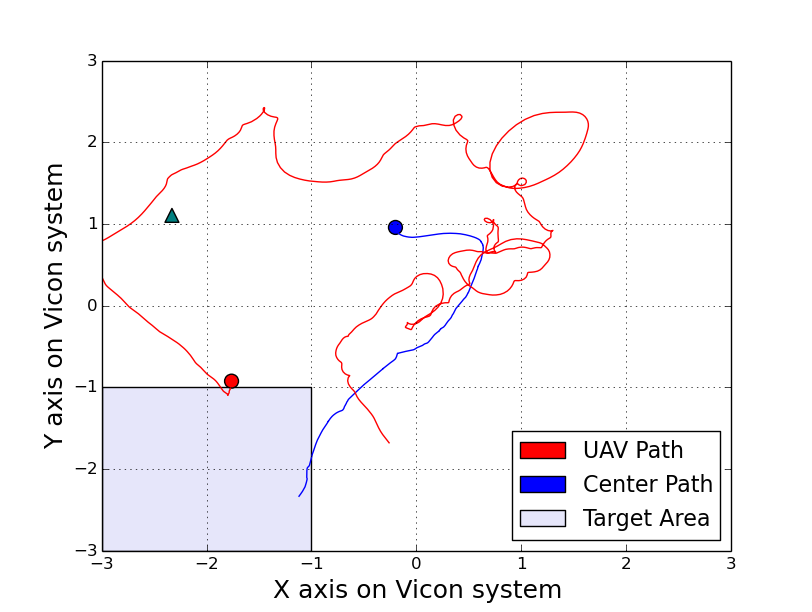}
        \label{fig:4x4to6x6-phy-traj}
    }\\
    \subfigure[HDDPG-6x6to6x6-Sim:Errors]
    {
        \includegraphics[width=0.25\textwidth]{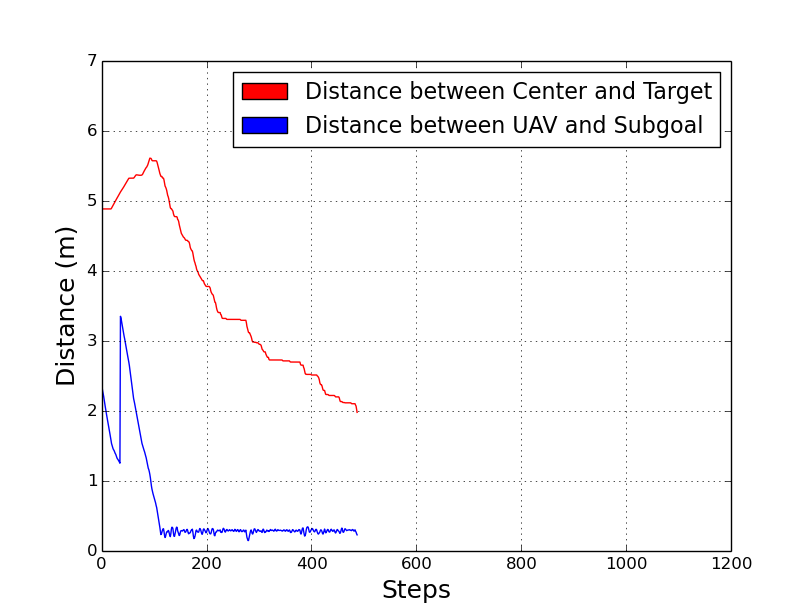}
        \label{fig:6x6to6x6-errors}
    }%
    \subfigure[HDDPG-4x4to6x6-Sim:Errors]
    {
        \includegraphics[width=0.25\textwidth]{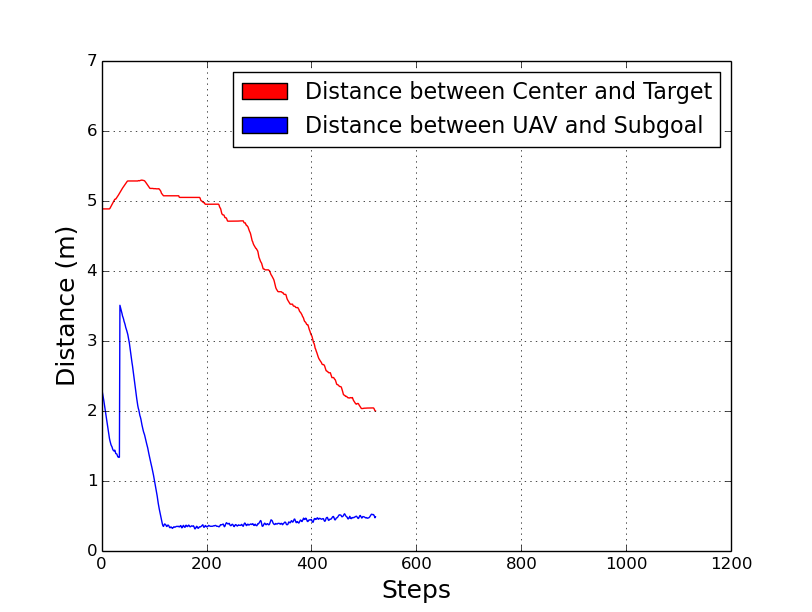}
        \label{fig:4x4to6x6-errors}
    }%
     \subfigure[HDDPG-6x6to6x6-Phy:Errors]
    {
        \includegraphics[width=0.25\textwidth]{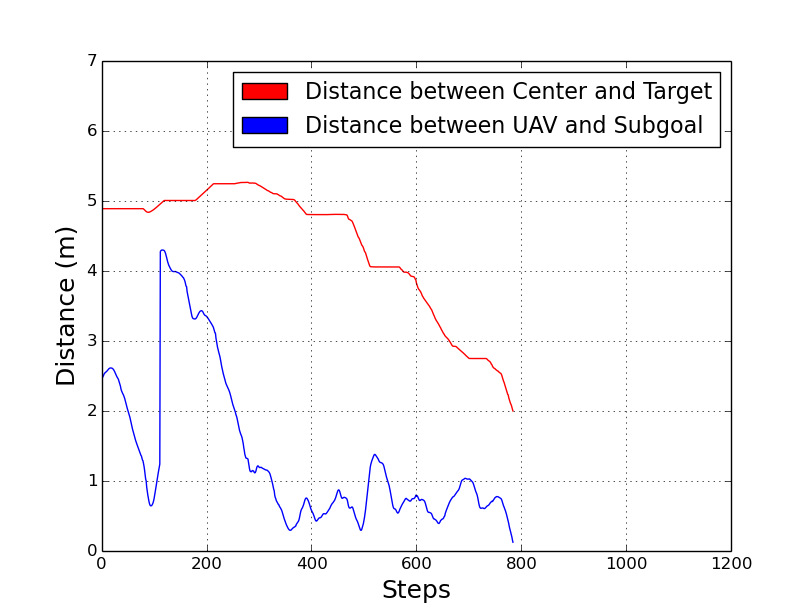}
        \label{fig:6x6to6x6-phy-errors}
    }%
    \subfigure[HDDPG-4x4to6x6-Phy:Errors]
    {
        \includegraphics[width=0.25\textwidth]{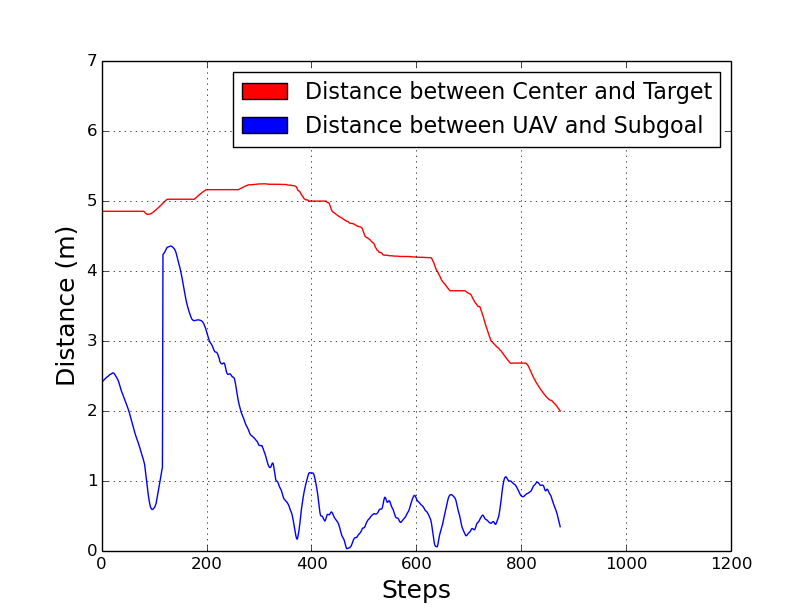}
        \label{fig:4x4to6x6-phy-errors}
    }
    \caption{Trajectories of the UAV and the center of mass and errors of the distance from the UAV to the sub-goal and the reduced distance from the center to the target in a testing case performed in a 6$\times$6 simulation and physical environment.}\label{fig:trajectories_phy}
\end{figure}

Similar behavior is observed in the physical environment. The travelled distance of the 4$\times$4 aerial shepherd is smaller than that of the 6$\times$6 agent, which are $24.1m$ and $27.6m$. Compared to this value of the agents in the simulation, both of them are higher. However, when we inspect the error per step of the agents, they are the same value of $0.05m$. This value is considerably higher than their errors per step in the simulation. This difference leads to the UAV agents needing to move longer to achieve the task in the physical environment. This is seen in the distances per step between the position of the UAV and the sub-goal. These distances cause the driving control of the UAV to the UGVs be undesirable. It is also understandable when the reduced distance from the center to target is smaller than that of the agents tested in the simulation so that the number of steps of the agents in the physical is higher than that in the simulation. However, it is worth noting that in the 6$\times$6 area of the physical environment, this error per step is acceptable~\cite{tran2019,tran2020switching}. 

Some general observations are shown in Figure~\ref{fig:trajectories_phy}. Although the paths of the 4$\times$4 and 6$\times$6 agent in the physical environment are longer and more oscillatory than that in the simulation, they are highly similar in the shape of the movement when the agents collect the furthest sheep, and then herd the entire swarm of the UGVs towards the target successfully. The error comes from the disturbances and natural offset behaviors of the drone in moving even though being stabilized by the position tracking system. Figure~\ref{fig:trajectories_phy} confirms that the distance from the UAV to the sub-goal in the simulation is stable and smaller than the corresponding distance in the physical environment. However, the distance between the center and the target of the agents in both the simulation and physical tends to non-monotonically decrease until the task is completed. 

From these testing results, we can see that the proposed learning approach shows promise for producing successful aerial shepherds. Although there are gaps between the performance of the agents in the simulation and the physical environment due to disturbances and un-modelled dynamics,
these gaps do not significantly change the behaviour. Additionally, within both the simulation and physical environments, we show that it is feasible that an agent trained in a smaller environment can transfer its skills to a larger environment.

\section{Conclusion and Future Work}\label{section-7}

In this paper, we have introduced a deep hierarchical reinforcement learning framework for decomposing a complex shepherding problem of ground-air vehicles into simpler sub-problem spaces and training the UAV agent to obtain the desired behavior in each case. The deep deterministic policy gradient (DDPG) networks demonstrate effective learning capabilities that achieve near-optimal solutions with dynamic, continuous environment and output continuous values as velocity vectors which are easy to execute through the control systems. The framework with the trained networks are then tested on the entire UAV-UGV shepherding mission where the objective is fulfilled with behavior emerged from combining low-level actions learned through interacting with two simpler search spaces.

The framework is tested in simulated and physical environments of the same size. We compared the H-DDPG approach against DHRL~\cite{nguyen2019deep} and the Str\"{o}mbom method~\cite{strombom2014solving} in the simulated environment. In the physical environment, there is more uncertainty due to the limitation of dynamic modelling in simulation, which results in some offsets between the desired and actual trajectories. The difference in performances between the simulation and physical environments is insignificant.

The scalability of the framework is also examined through transfer learning with state and action scaling from a smaller environment of $4m\times4m$ to a larger one of $6m\times6m$. Our results demonstrate that the transferred model achieves slightly similar completion time and travel distance while performing in a more stable manner than the model trained from the target environment. 

Future directions include the need to validate the performance and scalability of our proposed framework with different UGV scenarios. There is also an opportunity to account for obstacles in the environment by adding navigation and path planning skills to the behavioural hierarchy. Last, but not least, adding more UAVs and allowing for coordination skills in the hierarchy will generalize the problem to more realistic settings. 

\section*{Acknowledgment}
This material is based upon work supported by the Air Force Office of Scientific Research and the Office of Naval Research - Global (ONR-G).


\bibliography{nn}

\begin{thebibliography}{10}
\expandafter\ifx\csname url\endcsname\relax
  \def\url#1{\texttt{#1}}\fi
\expandafter\ifx\csname urlprefix\endcsname\relax\def\urlprefix{URL }\fi
\expandafter\ifx\csname href\endcsname\relax
  \def\href#1#2{#2} \def\path#1{#1}\fi

\bibitem{martinez2007motion}
S.~Martinez, J.~Cortes, F.~Bullo, Motion coordination with distributed
  information, IEEE Control Systems Magazine 27~(4) (2007) 75--88.

\bibitem{Chang2012Fuzzy}
Y.~{Chang}, C.~{Chang}, C.~{Chen}, C.~{Tao}, Fuzzy sliding-mode formation
  control for multirobot systems: Design and implementation, IEEE Transactions
  on Systems, Man, and Cybernetics, Part B (Cybernetics) 42~(2) (2012)
  444--457.
\newblock \href {https://doi.org/10.1109/TSMCB.2011.2167679}
  {\path{doi:10.1109/TSMCB.2011.2167679}}.

\bibitem{carelli2006centralized}
R.~Carelli, C.~De~la Cruz, F.~Roberti, Centralized formation control of
  non-holonomic mobile robots, Latin American applied research 36~(2) (2006)
  63--69.

\bibitem{oh2017bio}
H.~Oh, A.~R. Shirazi, C.~Sun, Y.~Jin, Bio-inspired self-organising multi-robot
  pattern formation: A review, Robotics and Autonomous Systems 91 (2017)
  83--100.

\bibitem{wen2018swarm}
J.~{Wen}, L.~{He}, F.~{Zhu}, Swarm robotics control and communications:
  Imminent challenges for next generation smart logistics, IEEE Communications
  Magazine 56~(7) (2018) 102--107.
\newblock \href {https://doi.org/10.1109/MCOM.2018.1700544}
  {\path{doi:10.1109/MCOM.2018.1700544}}.

\bibitem{yan2019efficient}
F.~Yan, K.~Di, J.~Jiang, Y.~Jiang, H.~Fan, Efficient decision-making for
  multiagent target searching and occupancy in an unknown environment, Robotics
  and Autonomous Systems 114 (2019) 41--56.

\bibitem{long2019comprehensive}
N.~K. Long, K.~Sammut, D.~Sgarioto, M.~Garratt, H.~Abbass, A comprehensive
  review of shepherding as a bio-inspired swarm-robotics guidance approach,
  arXiv preprint arXiv:1912.07796 (2019).

\bibitem{strombom2014solving}
D.~Str{\"o}mbom, R.~P. Mann, A.~M. Wilson, S.~Hailes, A.~J. Morton, D.~J.
  Sumpter, A.~J. King, Solving the shepherding problem: heuristics for herding
  autonomous, interacting agents, Journal of the royal society interface
  11~(100) (2014) 20140719.

\bibitem{Chen2016coordination}
J.~{Chen}, X.~{Zhang}, B.~{Xin}, H.~{Fang}, Coordination between unmanned
  aerial and ground vehicles: A taxonomy and optimization perspective, IEEE
  Transactions on Cybernetics 46~(4) (2016) 959--972.
\newblock \href {https://doi.org/10.1109/TCYB.2015.2418337}
  {\path{doi:10.1109/TCYB.2015.2418337}}.

\bibitem{Minaeian2016autonomous}
S.~{Minaeian}, J.~{Liu}, Y.~{Son}, Vision-based target detection and
  localization via a team of cooperative {UAV} and {UGVs}, IEEE Transactions on
  Systems, Man, and Cybernetics: Systems 46~(7) (2016) 1005--1016.
\newblock \href {https://doi.org/10.1109/TSMC.2015.2491878}
  {\path{doi:10.1109/TSMC.2015.2491878}}.

\bibitem{mathew2015planning}
N.~Mathew, S.~L. Smith, S.~L. Waslander, Planning paths for package delivery in
  heterogeneous multirobot teams, IEEE Transactions on Automation Science and
  Engineering 12~(4) (2015) 1298--1308.

\bibitem{mathews2019supervised}
N.~Mathews, A.~L. Christensen, A.~Stranieri, A.~Scheidler, M.~Dorigo,
  Supervised morphogenesis: Exploiting morphological flexibility of
  self-assembling multirobot systems through cooperation with aerial robots,
  Robotics and autonomous systems 112 (2019) 154--167.

\bibitem{nguyen2017supervised}
H.~T. Nguyen, M.~Garratt, L.~T. Bui, H.~Abbass, Supervised deep actor network
  for imitation learning in a ground-air {UAV-UGVs} coordination task, in: 2017
  IEEE Symposium Series on Computational Intelligence (SSCI), IEEE, 2017, pp.
  1--8.

\bibitem{chaimowicz2004aerial}
L.~Chaimowicz, V.~Kumar, Aerial shepherds: Coordination among {UAVs} and swarms
  of robots, in: In Proc. of DARS’04, Citeseer, 2004.

\bibitem{mnih2016asynchronous}
V.~Mnih, A.~P. Badia, M.~Mirza, A.~Graves, T.~Lillicrap, T.~Harley, D.~Silver,
  K.~Kavukcuoglu, Asynchronous methods for deep reinforcement learning, in:
  International conference on machine learning, 2016, pp. 1928--1937.

\bibitem{yang2018hierarchical}
Z.~Yang, K.~Merrick, L.~Jin, H.~A. Abbass, Hierarchical deep reinforcement
  learning for continuous action control, IEEE transactions on neural networks
  and learning systems~(99) (2018) 1--11.

\bibitem{yasuda2018collective}
T.~{Yasuda}, K.~{Ohkura}, Collective behavior acquisition of real robotic
  swarms using deep reinforcement learning, in: 2018 Second IEEE International
  Conference on Robotic Computing (IRC), 2018, pp. 179--180.
\newblock \href {https://doi.org/10.1109/IRC.2018.00038}
  {\path{doi:10.1109/IRC.2018.00038}}.

\bibitem{gee2019transparent}
A.~Gee, H.~Abbass, Transparent machine education of neural networks for swarm
  shepherding using curriculum design, arXiv preprint arXiv:1903.09297 (2019).

\bibitem{clayton2019machine}
N.~R. Clayton, H.~Abbass, Machine teaching in hierarchical genetic
  reinforcement learning: Curriculum design of reward functions for swarm
  shepherding, arXiv preprint arXiv:1901.00949 (2019).

\bibitem{nguyen2019deep}
H.~Nguyen, T.~Nguyen, M.~Garratt, K.~Kasmarik, S.~Anavatti, M.~Barlow,
  H.~Abbass, A deep hierarchical reinforcement learner for aerial shepherding
  of ground swarms, in: Neural Information Processing: 26th International
  Conference, ICONIP 2019, Sydney, Australia, November 14-18, 2017,
  Proceedings, Springer, 2019.

\bibitem{oh2015survey}
K.-K. Oh, M.-C. Park, H.-S. Ahn, A survey of multi-agent formation control,
  Automatica 53 (2015) 424--440.

\bibitem{balch1998behavior}
T.~Balch, R.~C. Arkin, Behavior-based formation control for multirobot teams,
  IEEE Transactions on robotics and automation 14~(6) (1998) 926--939.

\bibitem{vasarhelyi2014outdoor}
G.~{Vásárhelyi}, C.~{Virágh}, G.~{Somorjai}, N.~{Tarcai}, T.~{Szörényi},
  T.~{Nepusz}, T.~{Vicsek}, Outdoor flocking and formation flight with
  autonomous aerial robots, in: 2014 IEEE/RSJ International Conference on
  Intelligent Robots and Systems, 2014, pp. 3866--3873.
\newblock \href {https://doi.org/10.1109/IROS.2014.6943105}
  {\path{doi:10.1109/IROS.2014.6943105}}.

\bibitem{Ramazani2017Rigidity}
S.~{Ramazani}, R.~{Selmic}, M.~{de Queiroz}, Rigidity-based multiagent layered
  formation control, IEEE Transactions on Cybernetics 47~(8) (2017) 1902--1913.
\newblock \href {https://doi.org/10.1109/TCYB.2016.2568164}
  {\path{doi:10.1109/TCYB.2016.2568164}}.

\bibitem{sen2017cooperative}
A.~Sen, S.~R. Sahoo, M.~Kothari, Cooperative formation control strategy in
  heterogeneous network with bounded acceleration, in: Control Conference
  (ICC), 2017 Indian, IEEE, 2017, pp. 344--349.

\bibitem{guillet2017formation}
A.~Guillet, R.~Lenain, B.~Thuilot, V.~Rousseau, Formation control of
  agricultural mobile robots: A bidirectional weighted constraints approach,
  Journal of Field Robotics (2017).

\bibitem{miao2018distributed}
Z.~Miao, Y.-H. Liu, Y.~Wang, G.~Yi, R.~Fierro, Distributed estimation and
  control for leader-following formations of nonholonomic mobile robots, IEEE
  Transactions on Automation Science and Engineering 15~(4) (2018) 1946--1954.

\bibitem{Hung2017qlearning}
S.~{Hung}, S.~N. {Givigi}, A {Q}-learning approach to flocking with {UAVs} in a
  stochastic environment, IEEE Transactions on Cybernetics 47~(1) (2017)
  186--197.
\newblock \href {https://doi.org/10.1109/TCYB.2015.2509646}
  {\path{doi:10.1109/TCYB.2015.2509646}}.

\bibitem{yi2017bio}
X.~{Yi}, A.~{Zhu}, S.~X. {Yang}, C.~{Luo}, A bio-inspired approach to task
  assignment of swarm robots in 3-{D} dynamic environments, IEEE Transactions
  on Cybernetics 47~(4) (2017) 974--983.
\newblock \href {https://doi.org/10.1109/TCYB.2016.2535153}
  {\path{doi:10.1109/TCYB.2016.2535153}}.

\bibitem{santoso2017state}
F.~Santoso, M.~A. Garratt, S.~G. Anavatti, State-of-the-art intelligent flight
  control systems in unmanned aerial vehicles, IEEE Transactions on Automation
  Science and Engineering 15~(2) (2017) 613--627.

\bibitem{Yang2018leader}
Y.~{Yang}, H.~{Modares}, D.~C. {Wunsch}, Y.~{Yin}, Leader-follower output
  synchronization of linear heterogeneous systems with active leader using
  reinforcement learning, IEEE Transactions on Neural Networks and Learning
  Systems 29~(6) (2018) 2139--2153.
\newblock \href {https://doi.org/10.1109/TNNLS.2018.2803059}
  {\path{doi:10.1109/TNNLS.2018.2803059}}.

\bibitem{singh2016navigation}
P.~{Singh}, R.~{Tiwari}, M.~{Bhattacharya}, Navigation in multi robot system
  using cooperative learning: A survey, in: 2016 International Conference on
  Computational Techniques in Information and Communication Technologies
  (ICCTICT), 2016, pp. 145--150.
\newblock \href {https://doi.org/10.1109/ICCTICT.2016.7514569}
  {\path{doi:10.1109/ICCTICT.2016.7514569}}.

\bibitem{asada1999cooperative}
M.~Asada, E.~Uchibe, K.~Hosoda, Cooperative behavior acquisition for mobile
  robots in dynamically changing real worlds via vision-based reinforcement
  learning and development, Artificial Intelligence 110~(2) (1999) 275--292.

\bibitem{Zema2019formation}
N.~R. {Zema}, D.~{Quadri}, S.~{Martin}, O.~{Shrit}, Formation control of a
  mono-operated {UAV} fleet through ad-hoc communications: a {Q}-learning
  approach, in: 2019 16th Annual IEEE International Conference on Sensing,
  Communication, and Networking (SECON), 2019, pp. 1--6.
\newblock \href {https://doi.org/10.1109/SAHCN.2019.8824932}
  {\path{doi:10.1109/SAHCN.2019.8824932}}.

\bibitem{nguyen2018swarm}
T.~Nguyen, H.~Nguyen, E.~Debie, K.~Kasmarik, M.~Garratt, H.~Abbass, Swarm
  {Q}-learning with knowledge sharing within environments for formation
  control, in: 2018 International Joint Conference on Neural Networks (IJCNN),
  IEEE, 2018, pp. 1--8.

\bibitem{de2019bio}
J.~P. L.~S. de~Almeida, R.~T. Nakashima, F.~Neves-Jr, L.~V.~R. de~Arruda,
  Bio-inspired on-line path planner for cooperative exploration of unknown
  environment by a multi-robot system, Robotics and Autonomous Systems 112
  (2019) 32--48.

\bibitem{palmer2018lenient}
G.~Palmer, K.~Tuyls, D.~Bloembergen, R.~Savani, Lenient multi-agent deep
  reinforcement learning, in: Proceedings of the 17th International Conference
  on Autonomous Agents and MultiAgent Systems, International Foundation for
  Autonomous Agents and Multiagent Systems, 2018, pp. 443--451.

\bibitem{mnih2015human}
V.~Mnih, K.~Kavukcuoglu, D.~Silver, A.~A. Rusu, J.~Veness, M.~G. Bellemare,
  A.~Graves, M.~Riedmiller, A.~K. Fidjeland, G.~Ostrovski, et~al., Human-level
  control through deep reinforcement learning, Nature 518~(7540) (2015) 529.

\bibitem{lillicrap2015continuous}
T.~P. Lillicrap, J.~J. Hunt, A.~Pritzel, N.~Heess, T.~Erez, Y.~Tassa,
  D.~Silver, D.~Wierstra, Continuous control with deep reinforcement learning,
  arXiv preprint arXiv:1509.02971 (2015).

\bibitem{Gebhardt2018learning}
G.~H.~W. {Gebhardt}, K.~{Daun}, M.~{Schnaubelt}, G.~{Neumann}, Learning robust
  policies for object manipulation with robot swarms, in: 2018 IEEE
  International Conference on Robotics and Automation (ICRA), 2018, pp.
  7688--7695.
\newblock \href {https://doi.org/10.1109/ICRA.2018.8463215}
  {\path{doi:10.1109/ICRA.2018.8463215}}.

\bibitem{nguyen2018apprenticeship}
H.~Nguyen, M.~Garratt, H.~Abbass, Apprenticeship bootstrapping, in: 2018
  International Joint Conference on Neural Networks (IJCNN), IEEE, 2018, pp.
  1--8.

\bibitem{santana2015outdoor}
L.~V. Santana, A.~S. Brandao, M.~Sarcinelli-Filho, Outdoor waypoint navigation
  with the {AR Drone} quadrotor, in: 2015 International Conference on Unmanned
  Aircraft Systems (ICUAS), IEEE, 2015, pp. 303--311.

\bibitem{santana2014trajectory}
L.~V. Santana, A.~S. Brandao, M.~Sarcinelli-Filho, R.~Carelli, A trajectory
  tracking and {3D} positioning controller for the {AR Drone} quadrotor, in:
  2014 international conference on unmanned aircraft systems (ICUAS), IEEE,
  2014, pp. 756--767.

\bibitem{kingma2014adam}
D.~P. Kingma, J.~Ba, Adam: A method for stochastic optimization, arXiv preprint
  arXiv:1412.6980 (2014).

\bibitem{huang2014tum}
H.~Huang, J.~Sturm, Tum simulator, \url{http://wiki. ros. org/tum\_simulator},
  accessed: 2019-06-20 (2014).

\bibitem{tran2019}
P.~V. {Tran}, F.~{Santoso}, M.~A. {Garratt}, I.~R. {Petersen}, Adaptive second
  order strictly negative imaginary controllers based on the interval type-2
  fuzzy systems for a hovering quadrotor with uncertainties, IEEE/ASME
  Transactions on Mechatronics (2019) 1--1\href
  {https://doi.org/10.1109/TMECH.2019.2941525}
  {\path{doi:10.1109/TMECH.2019.2941525}}.

\bibitem{fernandez2017}
R.~A.~S. Fern{\'a}nde, S.~Dominguez, P.~Campoy, L 1 adaptive control for wind
  gust rejection in quad-rotor {UAV} wind turbine inspection, in: 2017
  International Conference on Unmanned Aircraft Systems (ICUAS), 2017.

\bibitem{tran2017}
V.~P. Tran, M.~Garratt, I.~R. Petersen, Formation control of multi-{UAV}s using
  negative-imaginary systems theory, in: 2017 11th Asian Control Conference
  (ASCC), IEEE, 2017, pp. 2031--2036.

\bibitem{tran2020switching}
V.~P. Tran, M.~Garratt, I.~R. Petersen, Switching time-invariant formation
  control of a collaborative multi-agent system using negative imaginary
  systems theory, Control Engineering Practice 95 (2020) 104245.

\end{thebibliography}

\noindent {\includegraphics[width=0.9in,clip,keepaspectratio]{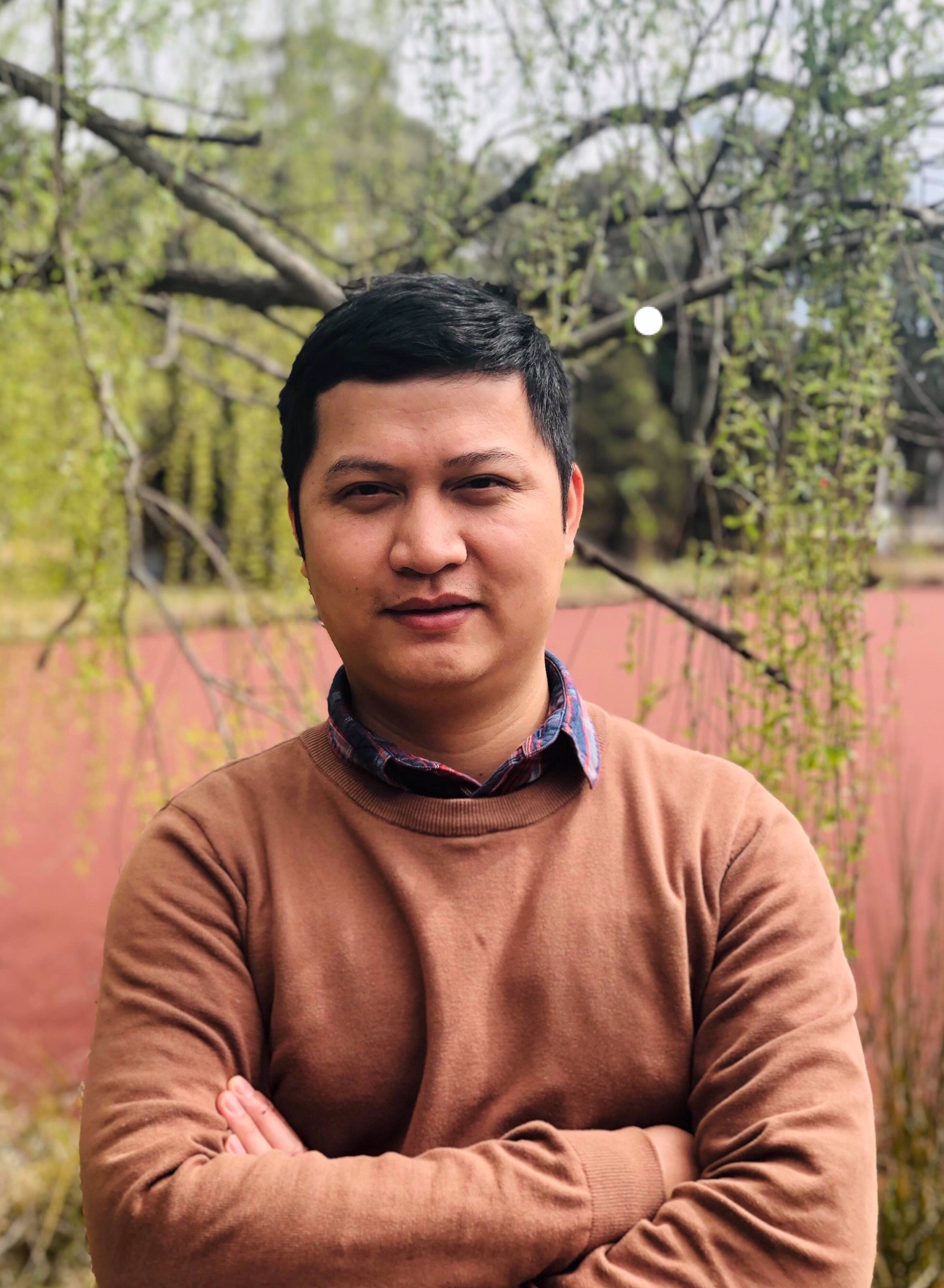}\textbf{Hung Nguyen} is pursuing a PhD at the University of New South Wales, Canberra, Australia. He received his B.E degree in Information Technology from Military Technical Academy, Vietnam in 2010 and his MSc in Computer Science from University of New South Wales - Canberra in 2018. His current research focuses on apprenticeship bootstrapping, deep learning, reinforcement learning, imitation learning, unmanned vehicles and human-machine teaming.}
\subsection*{  } 
\noindent {\includegraphics[width=0.9in,clip,keepaspectratio]{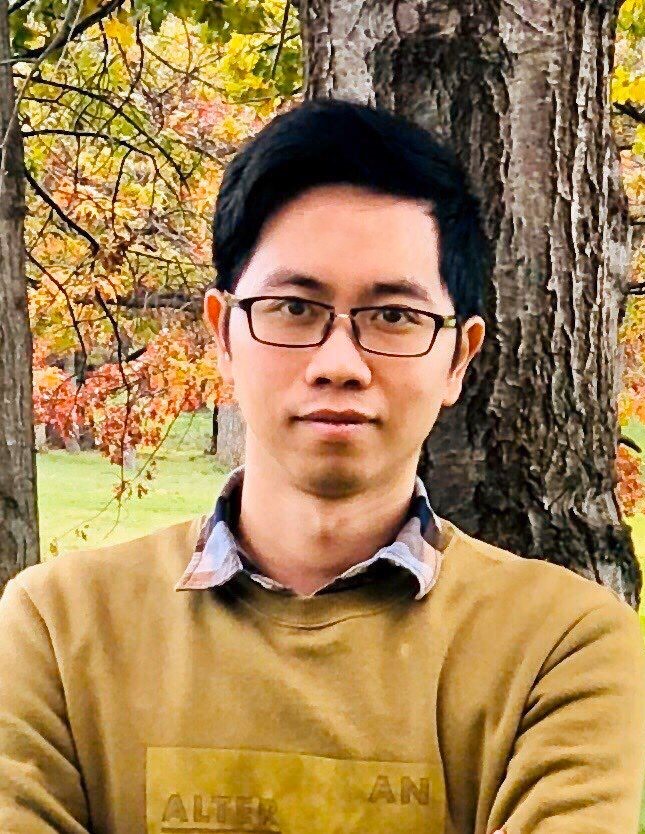}\textbf{Tung Duy Nguyen} is pursuing a PhD at the University of New South Wales, Canberra, Australia. He received his B.Eng. degree in Biomedical Engineering from Hanoi University of Science and Technology, Vietnam in 2015 and his MSc in Computer Science from University of New South Wales - Canberra in 2018. His research interests include deep learning, reinforcement learning, agent architecture, trusted autonomous systems, human-machine teaming, and interpretable and explainable artificial intelligence.}
\subsection*{  } 
\noindent {\includegraphics[width=0.9in,clip,keepaspectratio]{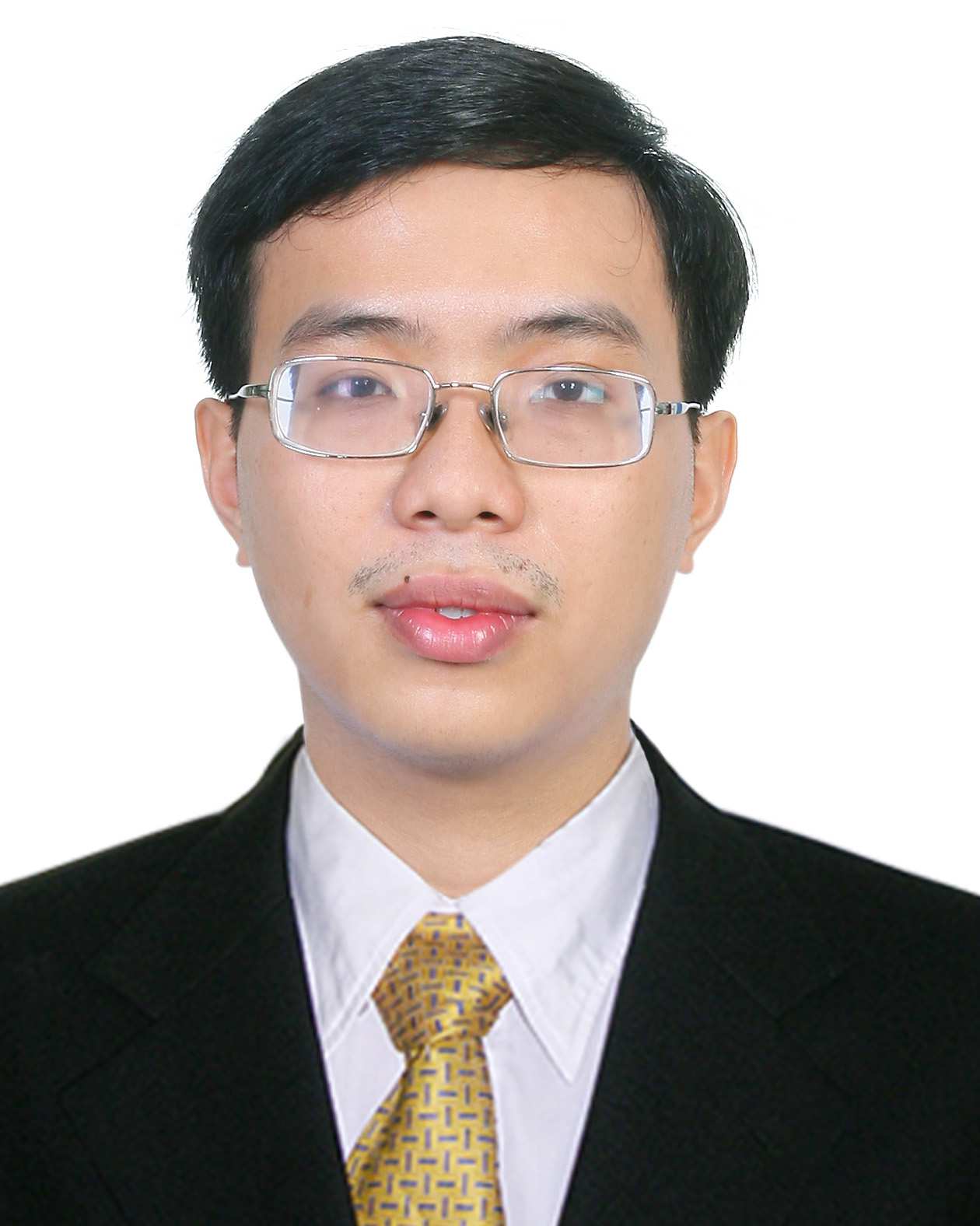}\textbf{Vu Phi Tran} received the B.E. degree in Automation and Control Engineering from the HCMC University of Technology and Education, Saigon, Vietnam, in 2010, the M.S. degree in Mechatronics Engineering from Asian Institute of Technology, Bangkok, Thailand, in 2015, and a PhD in the field of aerospace engineering from the University of New South Wales, Canberra, Australia, in 2019, where he is currently a Research Fellow. His research interests include adaptive control, robust control, consensus and formation control, neural networks, fuzzy systems, UAV systems and robotics.}
\subsection*{  } 
\noindent {\includegraphics[width=0.9in,clip,keepaspectratio]{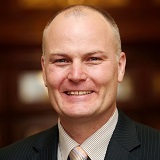}\textbf{Matthew Garratt} received a BE degree in Aeronautical Engineering from Sydney University, Australia, and a PhD in the field of biologically inspired robotics from the Australian National University. He is an Associate professor with UNSW-Canberra, Australia. His research interests include sensing, guidance and control for autonomous systems emphasising biologically inspired and computational intelligence approaches.}
\subsection*{  } 
\noindent {\includegraphics[width=0.9in,clip,keepaspectratio]{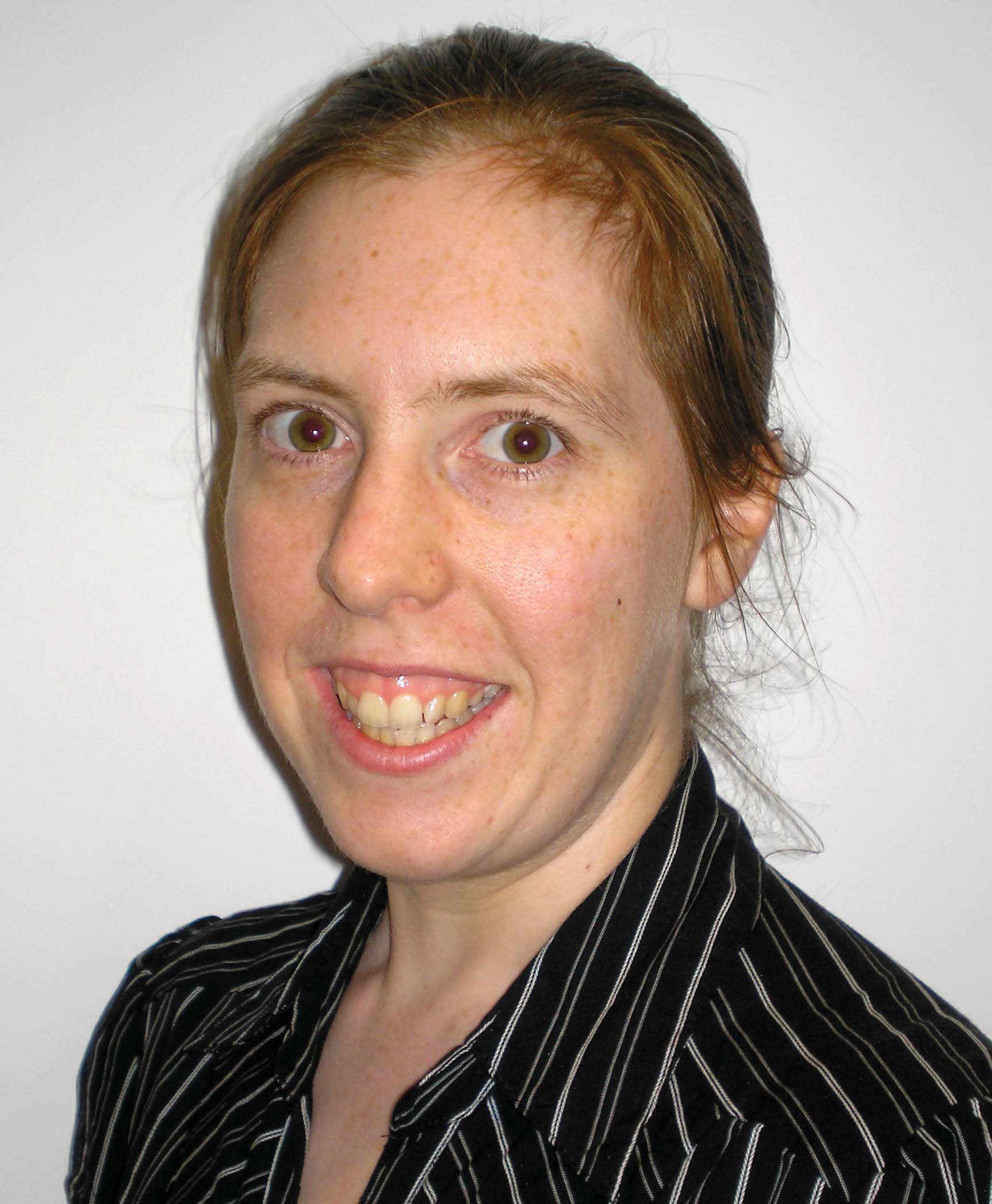}\textbf{Kathryn Kasmarik} Bachelor of Computer Science and Technology (Advanced, Honours I, University Medal), University of Sydney, Australia, 2002; PhD, University of Sydney, Australia, 2007. She is an Associate Professor in computer science at UNSW-Canberra, Australia. Her research lies in the area of autonomous mental development, with applications in computer games and developmental robotics.}

\subsection*{  } 
\noindent {\includegraphics[width=0.9in,clip,keepaspectratio]{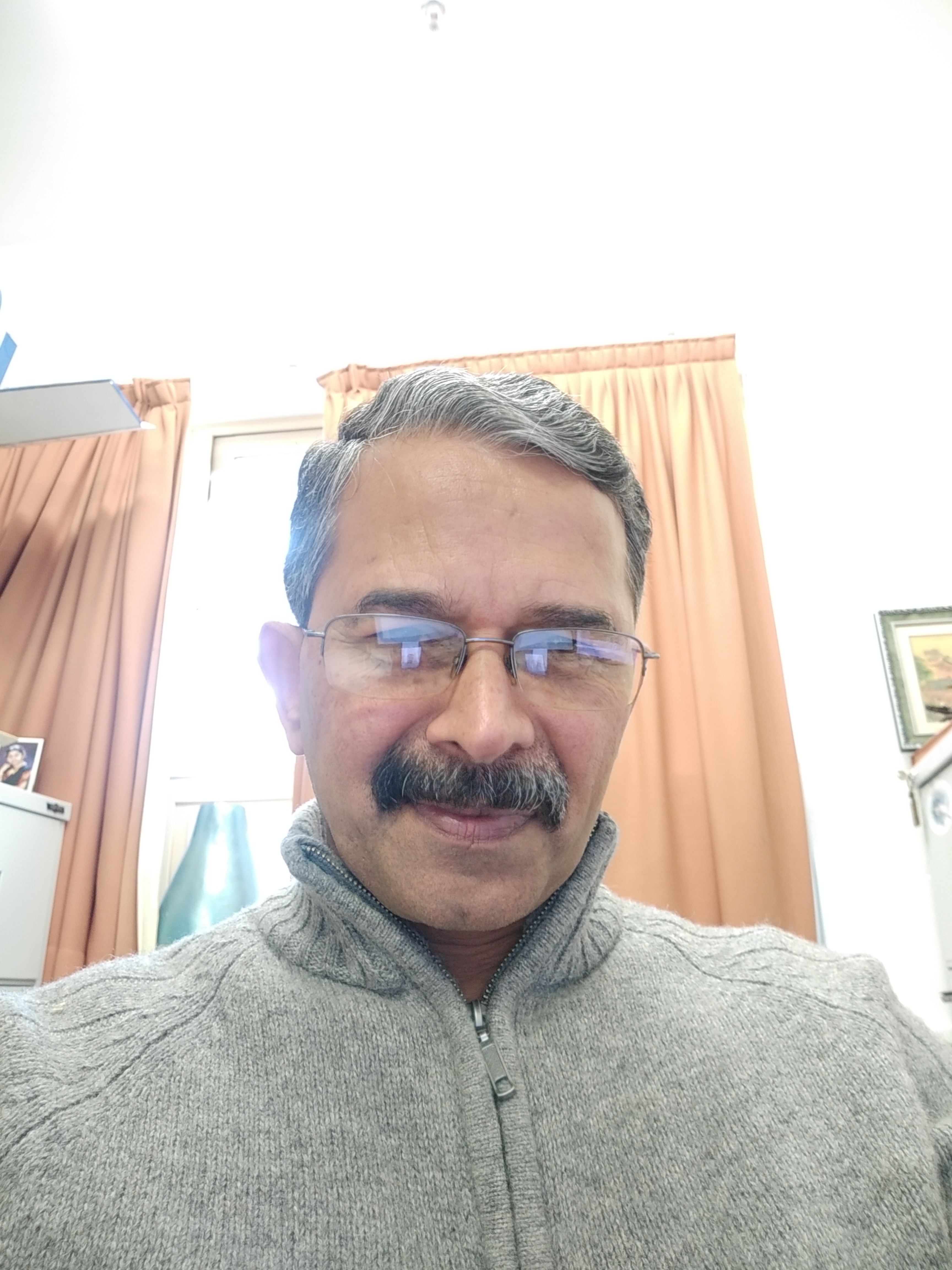}\textbf{Sreenatha Anavatti} is a senior lecturer with UNSW-Canberra. Before moving to Australia in 1998, he was an Associate Professor with the Indian Institute of Technology, Mumbai.  His research interests include the application of Fuzzy and Neural systems for UAVs, and Underwater and Ground Vehicles.}
\subsection*{  } 
\noindent {\includegraphics[width=0.9in,clip,keepaspectratio]{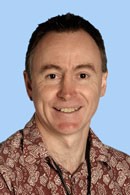}\textbf{Michael Barlow} received his PhD in Computer Science from UNSW-Canberra. Upon completion of two postdoctral researcher roles at the University of Queensland (Australia) and Nippon Telegraph and Telephone\textquoteright s Human Communication Laboratories in Japan, he joined UNSW Canberra in 1996. A/Prof. Michael Barlow's research interests include serious games, simulation, computational intelligence and HCI.}

\subsection*{  } 
\noindent {\includegraphics[width=0.9in,clip,keepaspectratio]{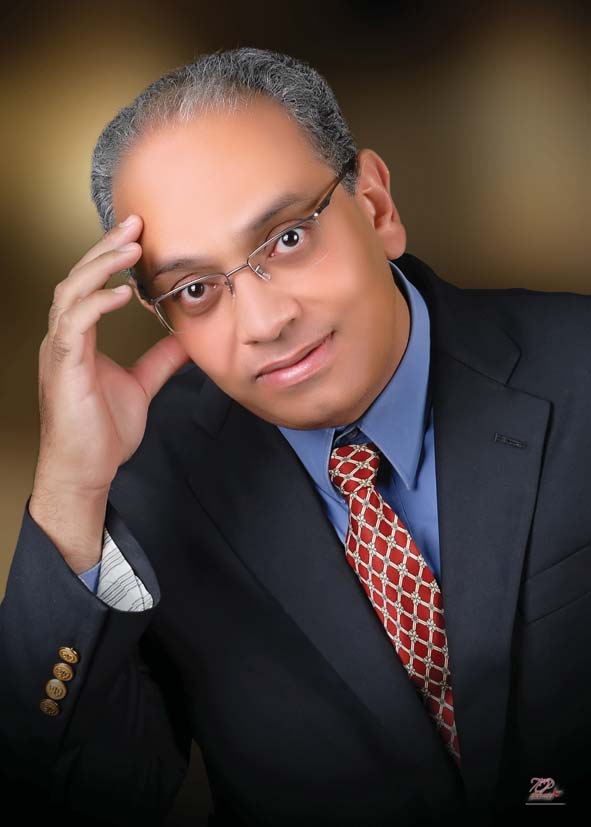}\textbf{Hussein A. Abbass} is a Professor at the University of New South Wales Canberra, Australia. He is the Founding Editor-in-Chief of the IEEE Transactions on Artificial Intelligence. He was the Vice-president for Technical
Activities (2016-2019) for the IEEE Computational Intelligence Society. His current research contributes to trusted human-swarm teaming with an aim to design next generation trusted and distributed artificial intelligence systems that seamlessly integrate humans and machines.}
\end{document}